%% file: main.tex
\documentclass[acmtog]{acmart}

\input{macros.tex}

\acmSubmissionID{279}

\usepackage{booktabs} 

\usepackage[ruled]{algorithm2e} 

\SetAlFnt{\small}
\SetAlCapFnt{\small}
\SetAlCapNameFnt{\small}
\SetAlCapHSkip{0pt}

\acmJournal{TOG}
\acmYear{2020}\acmVolume{39}\acmNumber{4}\acmArticle{76}\acmMonth{7} \acmDOI{10.1145/3386569.3392420}

\setcopyright{acmlicensed}

\begin{document}
\title{One Shot 3D Photography}

\author{Johannes Kopf}
\author{Kevin Matzen}
\author{Suhib Alsisan}
\author{Ocean Quigley}
\author{Francis Ge}
\author{Yangming Chong}
\author{Josh Patterson}
\author{Jan-Michael Frahm}
\author{Shu Wu}
\author{Matthew Yu}
\author{Peizhao Zhang}
\author{Zijian He}
\author{Peter Vajda}
\author{Ayush Saraf}
\author{Michael Cohen}
\affiliation{Facebook}

\renewcommand\shortauthors{Kopf et al.}

\begin{abstract}
\input{tex/abstract.tex}
\end{abstract}

%
%

%


\begin{teaserfigure}
\input{figures/teaser.tex}
\end{teaserfigure}
\maketitle

\input{tex/intro.tex}
\input{tex/previous.tex}

\input{tex/overview.tex}

\input{tex/depth.tex}
\input{tex/layering.tex}

\input{tex/inpainting.tex}
\input{tex/meshing.tex}
\input{tex/Viewing.tex}
\input{tex/results.tex}
\input{tex/conclusions.tex}

\bibliographystyle{ACM-Reference-Format}
\bibliography{bibliography}

\end{document}

%% file: macros.tex
\usepackage{enumitem}
\usepackage{soul}
\usepackage{xspace}
\usepackage{mathtools}
\usepackage{adjustbox}
\usepackage{colortbl}
\usepackage[binary-units=true]{siunitx}
\soulregister\xspace{0}

\DeclareRobustCommand{\unsure}[1]{{\sethlcolor{yellow}\hl{#1}}}

\newcommand{\johannes}[1]{{\color{magenta}\textbf{Johannes: }#1}\normalfont}
\newcommand{\kevin}[1]{{\color{green}\textbf{Kevin: }#1}\normalfont}

\newcommand{\rev}[1]{#1}

\long\def\ignorethis#1{}

\newcommand{\iitem}[1]{\item[\normalfont \textit{#1}]}

\newbox\jsavebox%
\newcommand{\jsubfig}[2]{%
	\sbox\jsavebox{\vspace{0mm}\centering#1}%
	\parbox[t]{\wd\jsavebox}{\vspace{0mm}\centering\usebox\jsavebox\\#2}%
	}

\def\naive{na{\"i}ve\xspace}

\def\x{\times}

\def\td{\tau_\textit{disp}}

%% file: tex/abstract.tex
3D photography is a new medium that allows viewers to more fully experience a captured moment.
In this work, we refer to a \emph{3D photo} as one that displays parallax induced by moving the viewpoint (as opposed to a stereo pair with a fixed viewpoint).
3D photos are static in time, like traditional photos, but are displayed with interactive parallax on mobile or desktop screens, as well as on Virtual Reality devices, where viewing it \emph{also} includes stereo.
We present an end-to-end system for creating and viewing 3D photos, and the algorithmic and design choices therein.
Our 3D photos are captured in a single shot and processed directly on a mobile device.
The method starts by estimating depth from the 2D input image using a new monocular depth estimation network that is optimized for mobile devices.
It performs competitively to the state-of-the-art, but has lower latency and peak memory consumption and uses an order of magnitude fewer parameters.
The resulting depth is lifted to a layered depth image, and new geometry is synthesized in parallax regions.
We synthesize color texture and structures in the parallax regions as well, using an inpainting network, also optimized for mobile devices, on the LDI directly.
Finally, we convert the result into a mesh-based representation that can be efficiently transmitted and rendered even on low-end devices and over poor network connections.
Altogether, the processing takes just a few seconds on a mobile device, and the result can be instantly viewed and shared.
We perform extensive quantitative evaluation to validate our system and compare its new components against the current state-of-the-art.
\ignorethis{
3D photography is a new medium that allows viewers to more fully experience a captured moment.
In this work, we refer to a \emph{3D photo} as one that displays parallax induced by moving the viewpoint motion (as opposed to a stereo pair with a predetermined fixed parallax).
3D photos are static in time, like traditional photos, but are displayed with interactive parallax on mobile or desktop screens, as well as on Virtual Reality devices, where viewing it \emph{also} includes stereo.
We present an end-to-end system for creating and viewing 3D photos, and the algorithmic and design choices therein.
\unsure{Captured in a single shot on a mobile device.}
\unsure{We present fast algorithms for processing image and depth, hallucinating occluded content, and creating a compact representation that can be efficiently transmitted and rendered even on low-end devices and network connections.}
\unsure{Depth estimation network --- high quality, optimized for mobile.}
\unsure{Inpainting network --- high quality, optimized for mobile.}
\unsure{Meshing --- ...}
\unsure{Quantitative, qualitative evaluation.}
}
\ignorethis{
3D Photos are a new medium that allows viewers to more fully experience a captured moment. Many thousands are being created and shared every hour. We present the algorithmic choices that went into the creation and presentation of 3D Photos. 3D Photos leverage a variety of modern computer vision and computer graphics technologies to efficiently process image+depth pairs to produce a compact and easy to share result. 3D Photos exhibit parallax through virtual camera motion that is mapped to scrolling and/or device motion. In VR 3D Photos are rendered in stereo and respond to head motion. To date, the depth has been derived from dual-camera phones. We also demonstrate new capabilities to leverage neural nets to determine depth from monocular images, thus extending the capability to many more devices, and even to apply to images from the past.}

%% file: figures/teaser.tex
\newlength\outerheight%
\newlength\innerheight%
\newlength\hgap%
\newlength\capgap%
\setlength\outerheight{4.2cm}%
\setlength\innerheight{3.7cm}%
\setlength\hgap{1.0mm}%
\setlength\capgap{-0.65mm}%
\centering%
\jsubfig{\vspace{0mm}\includegraphics[height=\outerheight]{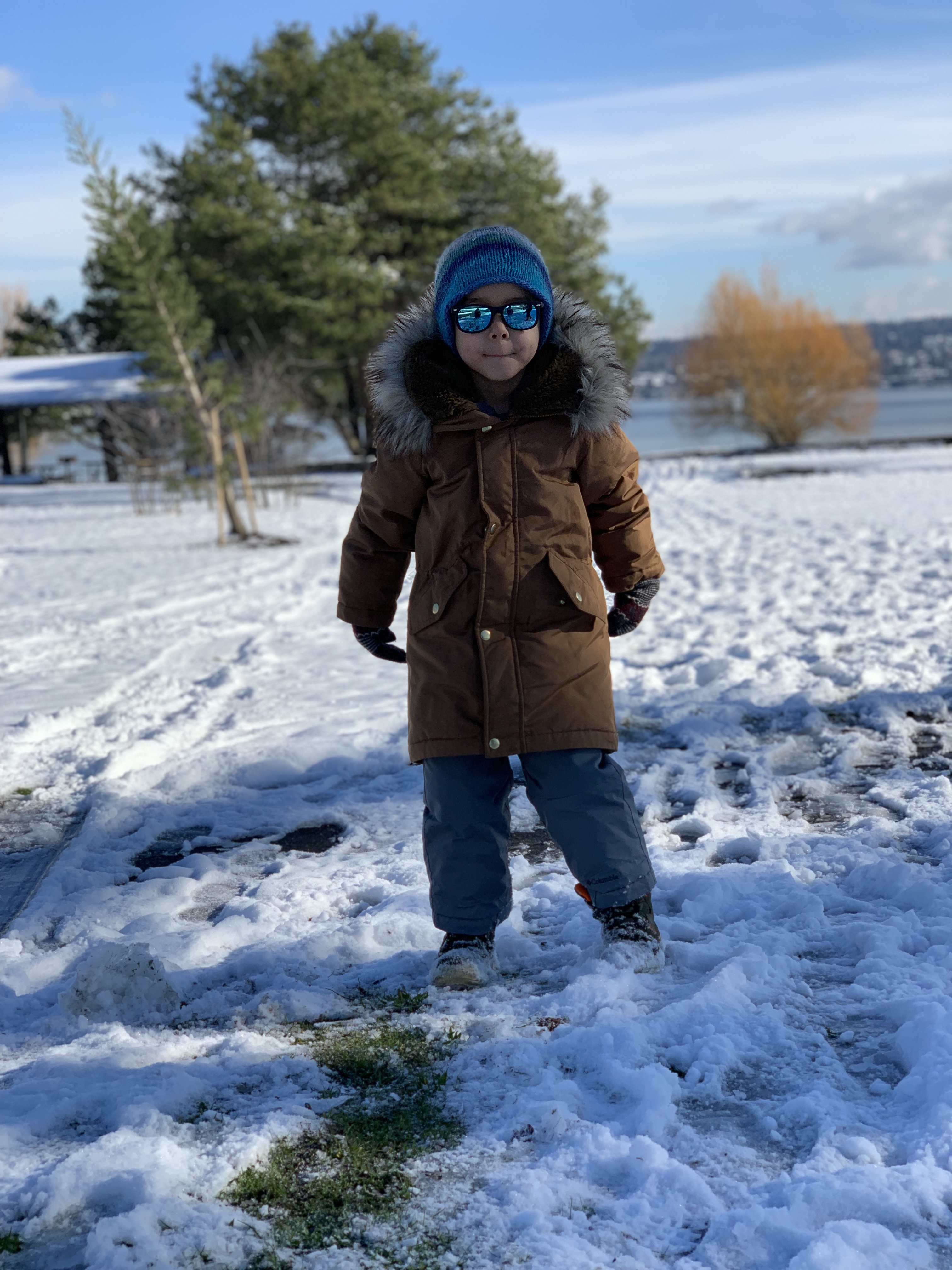}}{\vspace{\capgap}%
  \small (a) Input}%
\hfill%
$\underbracket[1pt][2.0mm]{%
\jsubfig{\vspace{0mm}\centering\includegraphics[height=\innerheight]{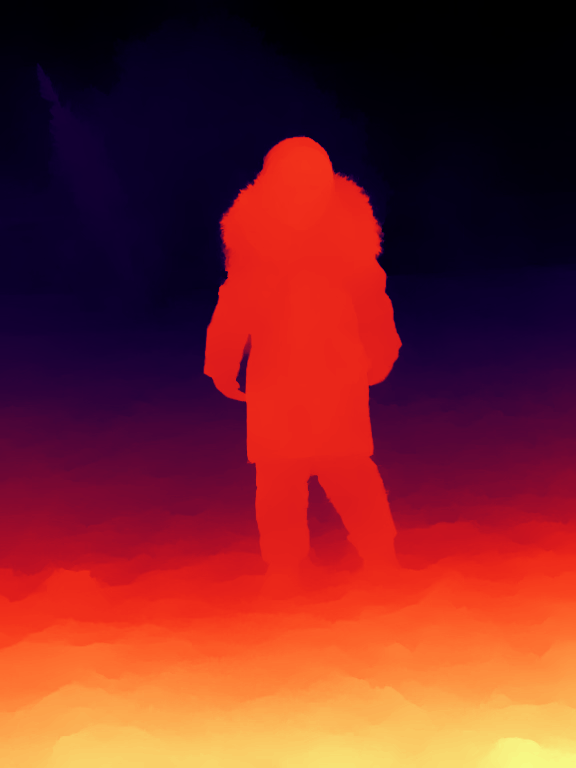}}{\vspace{\capgap}%
  \small (b) Depth estimation\\\footnotesize(230 ms)}%
\hspace{\hgap}%
\jsubfig{\vspace{0mm}\fbox{\includegraphics[height=\innerheight,trim=150 0 250 0,clip]{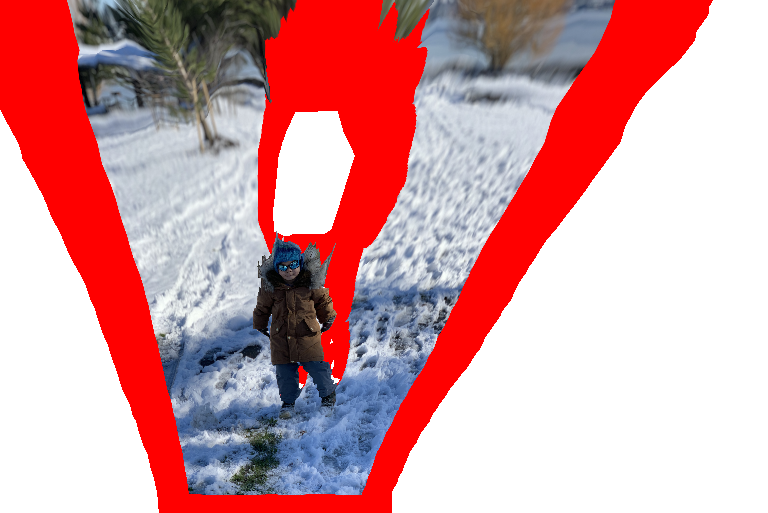}}}{\vspace{\capgap}%
  \small (c) Layer generation\\\footnotesize(94 ms)}%
\hspace{\hgap}%
\jsubfig{\vspace{0mm}\fbox{\includegraphics[height=\innerheight,trim=150 0 250 0,clip]{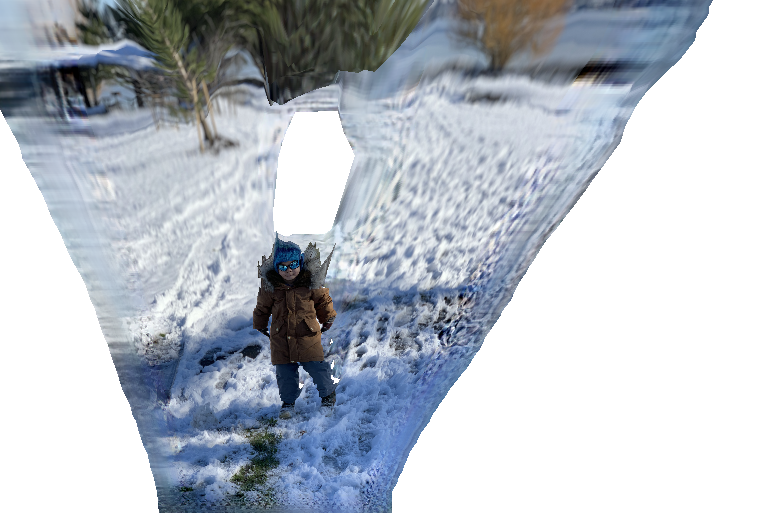}}}{\vspace{\capgap}%
  \small (d) Color inpainting\\\footnotesize(540 ms)}%
\hspace{\hgap}%
\jsubfig{\vspace{0mm}\includegraphics[height=\innerheight]{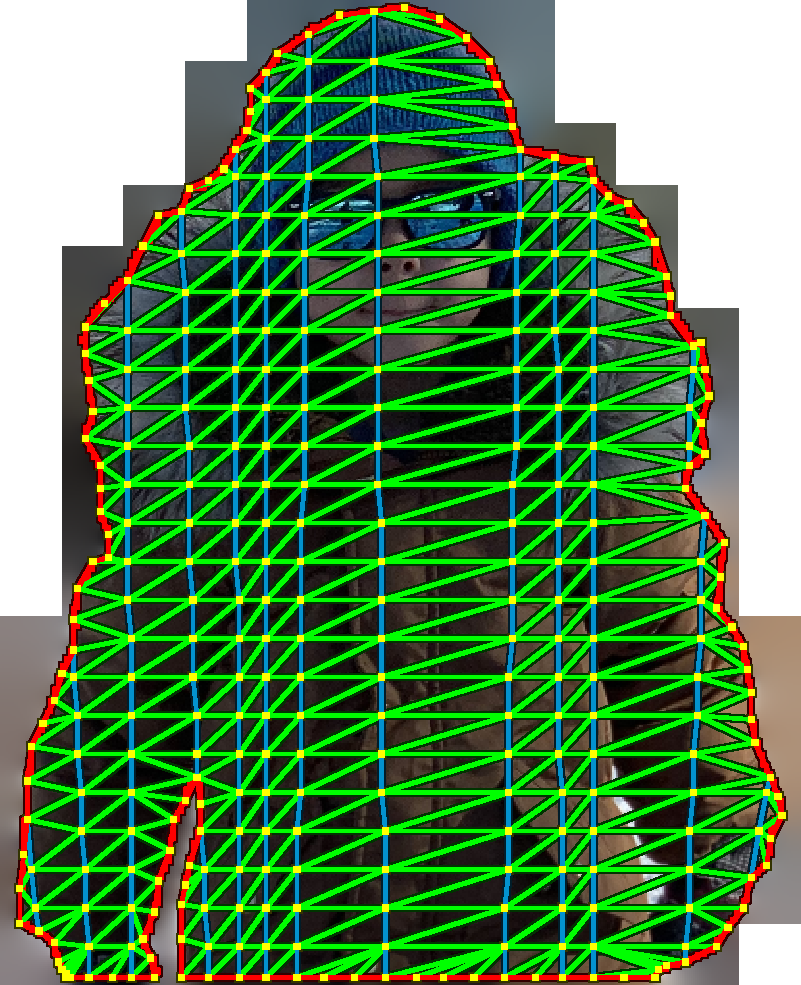}}{\vspace{\capgap}%
  \small (e) Meshing\\\footnotesize(234 ms)}%
 }_%
    {\substack{\vspace{-3.0mm}\\\colorbox{white}{~~Processing: 1,098ms on a mobile phone (iPhone 11 Pro)~~}}}$%
\hfill%
\jsubfig{\vspace{0mm}\includegraphics[height=\outerheight]{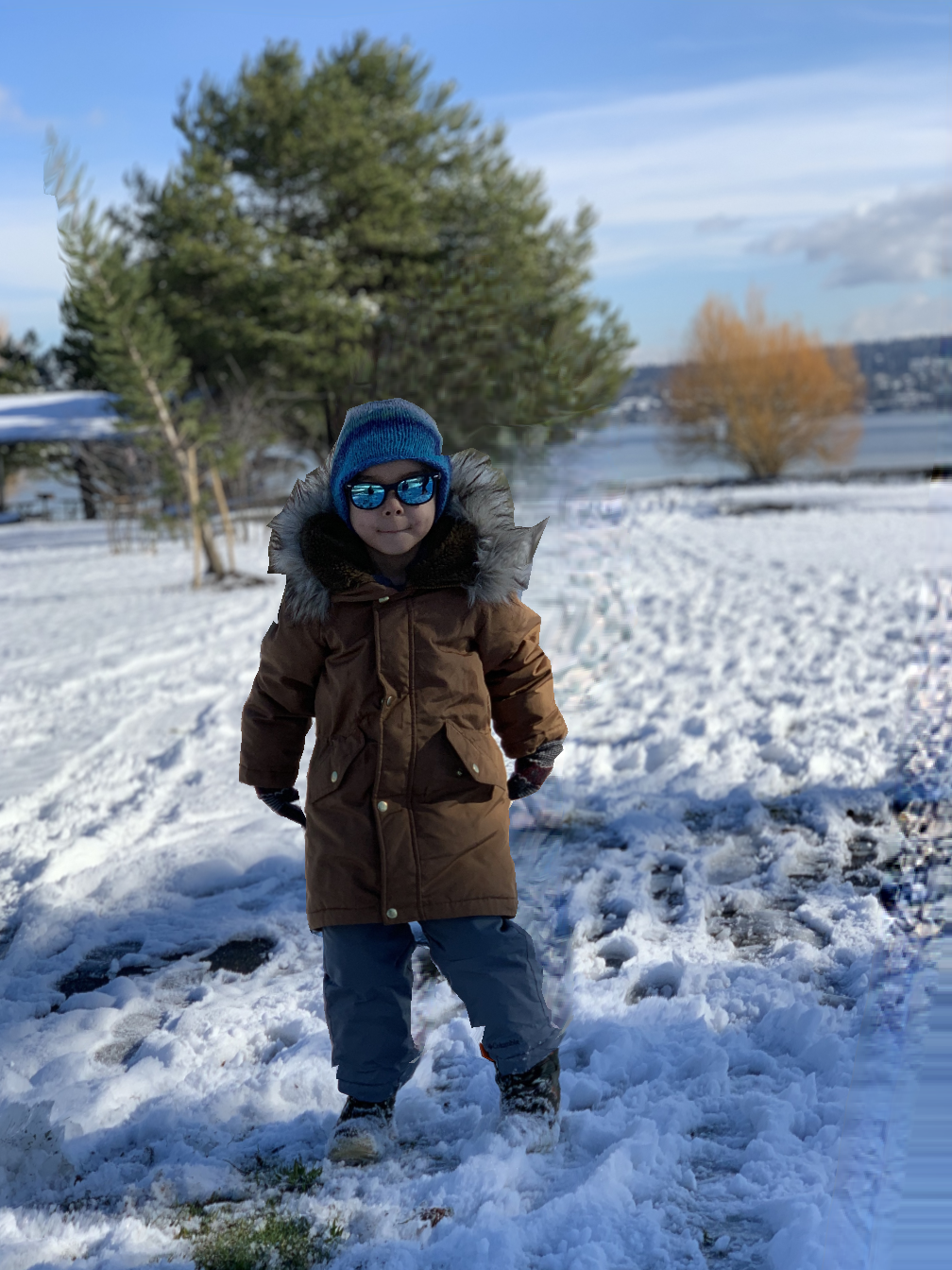}}{\vspace{\capgap}%
  \small (f) Novel view\\\footnotesize(real-time)}%
\vspace{-0mm}\\%
\def\teasercaption{
We present a system for creating \emph{3D photos} from a single mobile phone picture (a).
The process involves learning-based algorithms for estimating depth from the 2D input (b) and texture inpainting (d), as well as conventional algorithms for lifting the geometry to 3D and extending it in parallax regions (c), as well as generating a final mesh-based representation (e).
All steps are optimized to be fast given the limited compute and memory resources available on a mobile device.
The resulting representation (f) can be viewed instantly, generating novel viewpoints at real-time rates.
}%
\caption{\teasercaption}
\Description[We present a system for creating \emph{3D photos} from single-shot cell phone pictures.]{\teasercaption}
\undef\teasercaption
\label{fig:teaser}
\undef\outerheight%
\undef\innerheight%
\undef\hgap%
\undef\capgap%

%% file: tex/intro.tex
\section{Introduction}
\label{sec:intro}

Traditional 2D photography lets us capture the world around us, with a single click, as an instant frozen in time.
\emph{3D photography} is a new way to make these captured moments come back alive.
We use the term \emph{3D photo} to refer to any representation that can be displayed with parallax induced by viewpoint motion at viewing time (as opposed to a stereo pair, where inter-ocular parallax is baked in at capture time).
Although still static in time, 3D photos can be interactively explored.
The ability to change the viewpoint is compelling on ``flat'' mobile or desktop screens, and enables truly life-like experiences in Virtual Reality, by adding stereo viewing to head-motion induced parallax.

However, creating and displaying 3D photos poses challenges that are not present in 2D or even stereo photography:
dense depth is required in addition to color, viewpoint changes reveal previously occluded parts of the scene that must be filled, and the affordances for changing the viewpoint must be developed.

Accurately \rev{\emph{triangulating}} depth requires capturing at least two views of the scene;
reconstructing \emph{occluded} content requires even more captured views \cite{Hedman2017,Zitnick2004}.
This goes beyond the effort that most people are willing to spend on a photograph.
While some high-end smartphones are equipped with multi-lens cameras that can be used to estimate depth from stereo, they do not help with occluded regions, due to the small baseline of the lenses.
More importantly, there is currently at least an order of magnitude more phones in use that only have regular single-lens cameras.

We propose a system that provides a more practical approach to 3D photography.
Specifically, we address these design objectives:
\begin{description}[leftmargin=1.0\parindent,labelindent=0.0\parindent]
\iitem{Effort:} the capture should occur in a \emph{single shot} and not require any special hardware.
\iitem{Accessibility:} creation should be accessible on any mobile device, even devices with regular, single-lens cameras.
\iitem{Speed:} all post-capture processing should at most take a few seconds (on the mobile device) before the 3D photo can be viewed and shared.
\iitem{Compactness:} the final representation should be easy to transmit and display on low-end devices for sharing over the internet.
\iitem{Quality:} rendered novel views should look realistic; in particular, depth discontinuities and disocclusions should be handled gracefully.
\iitem{Intuitive Interaction:} \rev{interacting with a 3D photo must be in real-time, and the navigation affordances intuitive.}
\end{description}

Our system relies only on a single image as input, and estimates the depth of the scene as well as the content of parallax regions using learning-based methods.
It comprises four algorithm stages (Figure~\ref{fig:teaser}b--e), each containing new technical contributions:
\begin{description}[leftmargin=1.0\parindent,labelindent=0.0\parindent]
\iitem{Depth Estimation:} A dense depth map is estimated from the input image using a new neural network, constructed with efficient building blocks and optimized with automatic architecture search and \texttt{int8}-quantization for fast inference on mobile devices.
It performs competitively w.r.t. the state-of-the-art while consuming considerably fewer resources and having fewer parameters.
\iitem{Layer Generation:} The pixels are lifted onto a layered depth image (LDI), and we synthesize new geometry in parallax regions using carefully designed heuristic algorithms.
\iitem{Color Inpainting:} We synthesize colors for the newly synthesized geometry of the LDI using an inpainting neural network.  A novel set of neural modules enables us to transform this 2D CNN to one that can be applied to the LDI structure directly.
\iitem{Meshing:} Finally, we create a compact representation that can be efficiently rendered even on low-end devices and effectively transferred over poor network connections.
\end{description}

All processing steps are optimized for running fast on a mobile device with limited available resources. 
We also discuss affordences for viewing 3D photos on mobile and fixed flat screens, as well as using head-mounted displays for virtual reality.

We validate our system through extensive quantitative evaluation of our system's components, in particular, comparing depth estimation and inpainting to alternative state-or-the-art algorithms.
We believe that our proposed system, altogether, largely achieves the stated objectives and makes 3D photography truly practical and accessible for everyone.

%% file: tex/previous.tex

\section{Previous Work}
\label{sec:previous}

\paragraph{Manual annotation:}
The classic way to create 3D images, proposed in the seminal ``tours into the picture''~\cite{Horry1997}, involves carefully annotating the depth of a picture manually.
This can also be done semi-manually with the help of tools \cite{Oh2001}.
While the manual-assisted approach promises the ability to generate arbitrarily high quality depth maps for downstream processing, depth annotation is a laborious process and requires a skilled user.

\paragraph{Single-image depth estimation:}
Since the seminal paper of Saxena et al.~\shortcite{Saxena2006} there has been considerable progress on estimating depth from a single image, by leveraging advances in deep learning.
Chen et al.~\shortcite{Chen2016} propose a convolutional neural network (CNN) architecture and large quantities of training photos with ordinally labeled point pairs, and provide substantially improved generalization capability compared to previous work.
Li and Snavely~\shortcite{Li2018} provide even denser depth annotation from large-scale photometric reconstruction.
Networks can also be trained with a photometric loss and stereo supervision \cite{Garg2016,Godard2017,Kuznietsov2017,Godard2019}, which might be easier to obtain than depth annotation.
In addition, synthetic data~\cite{Ramamonjisoa2019,Niklaus2019} might help with synthesizing sharper depth discontinuities.
Ranftl et al.~\shortcite{Ranftl2019} show a good improvement by training from several datasets.
While the work mentioned above achieves commendable results, the proposed network architectures are too resource intensive in terms of processing, memory consumption and model size for mobile devices (Section~\ref{sec:depth_eval}).
We propose a new architecture in this work that performs competitively, but is considerably faster and smaller.

\rev{In terms of accelerating CNNs for monocular depth inference, Wofk et al.~\shortcite{Wofk2019}, Poggi et al.~\shortcite{Poggi2018}, and Peluso et al.~\shortcite{Peluso2019} each proposed a low-latency architecture for real-time processing on embedded platforms.
Lin et al.~\shortcite{Lin2020} explored reducing the memory footprint of monocular depth estimation networks by super-resolving predicted depth maps.
Finally, Tonioni et al.~\shortcite{Tonioni2019} proposed an online domain adaptation learning technique suitable for realtime stereo inference.
We compare against some of these methods in Section~\ref{sec:depth_eval}.}

\paragraph{Layered Depth Images:}
Shade et al.~\shortcite{Shade1998} provide a taxonomy of representations for 3D rendering and our work leverages one of them for processing.
In particular, we leverage \emph{Layered Depth Images} (LDI), similar to recent work~\cite{Hedman2017,Hedman2018}, but with more sophisticated heuristics for inpainting occlusions, and optimized algorithms to compute the result 
within seconds on mobile devices.
LDI provide an easy-to-use representation for background expansion and inpainting, and lend themselves for conversion into a textured triangle mesh for final content delivery and rendering.

\paragraph{Multi-plane Images:}
Stereo Magnification \cite{Zhou2018} proposed synthesizing a \emph{Multi-plane Image} (MPI) representation, i.e., a stack of fronto-parallel planes with RGB$\alpha$ textures, from a small-baseline stereo pair.
This work is extended to Srinivasan et al.~\shortcite{Srinivasan2019} to reduce the redundancy in the representation and expand the ability to change the viewpoint.
Flynn et al.~\shortcite{Flynn2019} generate high-quality MPIs from a handful of input views using learned gradient descent, and Mildenhall et al.~\shortcite{Mildenhall2019} blend a stack of MPIs at runtime.
All MPI generation methods above have in common that they require two or more views as input, while our proposed method uses only a single input image.

\paragraph{Other Representations and Neural Rendering:}
Sitzmann et al.~\shortcite{Sitzmann2019} encode the view-dependent appearance of a scene in a voxel grid of features and decode at runtime using a ``neural renderer''.
Other methods \cite{MartinBrualla2018,Meshry2019} also leverage on-the-fly neural rendering to increase the photorealism of their results.
However, these methods do not guarantee that disocclusions are filled consistently from different viewing angles, and they require too powerful hardware at runtime to perform in real-time on mobile devices.

\paragraph{Single-image Novel View Synthesis:}
Most aforementioned view synthesis methods require multiple input images, while there are only a few that can operate with a single input image, like ours.
This is important to mention, because view synthesis from a single input image is a considerably more difficult and ill-posed problem.
Yet, it is desirable, because requiring a user to capture a single view is more practical and the method can even be applied retro-actively to any existing photo, as demonstrated with historical photos in the accompanying video.

Liu et al.~\shortcite{Liu2018CVPR} predict a set of homography warps, and a selection map to combine the candidate images to a novel view.
It employs complex networks at runtime, leading to slow synthesis.
Srinivasan et al.~\shortcite{Srinivasan2017} predict a 4D light field representation.
This work has only been demonstrated in the context of narrow datasets (e.g., of plants, toys) and has not been shown to generalize to more diverse sets.

%% file: tex/overview.tex
\section{Overview}

3D photography requires a geometric representation of the scene.
There are many popular choices, although some have disadvantages for our application.
\emph{Light fields} capture very realistic scene appearance, but have excessive storage, memory, and processing requirements.
\emph{Meshes} and \emph{voxels} are very general representations, but are not optimized for being viewed from a particular viewpoint.
\emph{Multi-plane images} are not storage and memory efficient, and exhibit artifacts for sloped surfaces at large extrapolations.

In this paper we build on the \emph{Layered Depth Image (LDI)} representation \cite{Shade1998}, as in previous work \cite{Hedman2017,Hedman2018}.
An LDI consists of a regular rectangular lattice with integer coordinates, just like a normal image; but every position can hold zero, one, or more pixels.
Every LDI-pixel stores a color and a depth value.
Similar to Zitnick et al.~\shortcite{Zitnick2004}, we explicitly represent the 4-connectivity of pixels between and among layers, i.e., every pixel can have either zero or exactly one neighbor in each of the cardinal directions (left, right, up, down).

This representation has significant advantages:
\begin{description}
\iitem{Sparsity:} It only stores features that are present in the scene.
\iitem{Topology:} LDIs are locally like images. Many fast image processing algorithms translate to LDIs.
\iitem{Level-of-detail:} The regular sampling in image-space provides inherent level-of-detail: near geometry is more densely sampled than far geometry.
\iitem{Meshing:} LDIs can be efficiently converted into textured meshes~(Sections~\ref{sec:atlas}-\ref{sec:meshing}), which can be efficiently transmitted and rendered.
\end{description}

While LDIs have been used before to represent captured scenes \cite{Hedman2017,Hedman2018}, our work makes several important contributions:
(1) unlike previous work, our algorithm is not limited to only producing at most two layers at any point;
(2) we better shape the continuation of depth discontinuities into the disoccluded region using constraints;
(3) we propose a new network for inpainting occluded LDI pixels, as well as a method to translate existing 2D inpainting networks to operate directly on LDIs;
(4) efficient algorithms for creating texture atlases and simplified triangle meshes;
(5) our complete algorithm is faster and runs end-to-end in just a few seconds on a mobile device.

In the next section, we describe our algorithm for creating 3D photos from single color images.
Next, we describe in Section~\ref{sec:viewing} how they are experienced on mobile and fixed \emph{flat} screens, as well as using head-mounted displays for virtual reality.
Finally, in Section~\ref{sec:results} we provide detailed quantitative evaluation of our algorithm components as well as comparisons to other state-of-the-art methods.

%% file: tex/depth.tex
\section{Creating 3D Photos}

The input to our method is a single color image.
It is typically captured with a mobile phone, but any other photo may be used (e.g., historical pictures).

Our system comprises four stages (Figure~\ref{fig:teaser}b--e) and runs end-to-end on the mobile capture device.
We describe \emph{depth estimation} in Section~\ref{sec:depth}, lifting to an LDI and synthesizing occluded geometry in Section~\ref{sec:layering}, inpainting color on the occluded layers in Section~\ref{sec:inpainting}, and converting the LDI into the final mesh representation in Section~\ref{sec:meshing}.


\subsection{Depth Estimation}
\label{sec:depth}

\input{figures/tiefenrausch_scheme.tex}

The first step in our algorithm is to estimate a dense depth map from the input image.
Monocular depth estimation is a very active field, and many competitive methods have just appeared in the months prior to writing \cite{Ranftl2019,Ramamonjisoa2019,Niklaus2019,Godard2019}.
While these methods achieve high quality results, they use large models that consume considerable resources during inference.
This makes it difficult to deploy them in a mobile application.
In fact, most of these methods cannot run even on high-end smart phones due to the limited memory on these platforms (see Section~\ref{sec:depth_eval}).

In this section we propose a new architecture, called \emph{Tiefenrausch}, that is optimized to consume considerably fewer resources, as measured in terms of inference latency, peak memory consumption, and model size, while still performing competitively to the state-of-the-art.

These improvements were achieved by combining three techniques:
(1) building an \emph{efficient block structure} that is fast on mobile devices,
(2) using a \emph{neural architecture search} algorithm to find a network design that achieves a more favorable trade-off between accuracy, latency, and model size, and, then,
(3) using 8-bit \emph{quantization} to achieve a further reduction of the model size and latency while retaining most of the accuracy.
Below, we describe these optimizations as well as the training procedure in detail.

\paragraph{Efficient Block Structure}
We built an efficient block structure inspired by previous work \cite{Sandler2018, Wu2018} and is illustrated in Fig. ~\ref{fig:block_structure}.
The block contains a sequence of point-wise (1x1) convolution, KxK depthwise convolution where K is the kernel size, and another point-wise convolution.
Channel expansion, $e$, is a multiplicative factor which increases the number of channels after the initial point-wise convolution.
In layers which decrease the spatial resolution, depthwise convolution with stride, $s_d > 1$, is used.
When increasing the spatial resolution, we use nearest neighbor interpolation with a scale factor, $s_u > 1$, after the initial point-wise convolution.
If the output dimensions of the block are the same as the input dimensions (i.e., $s_d=s_u=1$, $C_{in}=C_{out}$), then a skip connection is added between the input and output with an additional block in the middle.

We combine these blocks into a U-Net like architecture \cite{Chen2016, Li2018, Ronneberger2015} as shown in Fig. ~\ref{fig:tr_scheme}.
We fixed the number of downsampling stages to 5 where each stage has a downsampling factor $s_d=2$. 
All stages have 3 blocks per stage and skip connections are placed between stages with the same spatial resolution.

\input{figures/block_structure.tex}

\paragraph{Neural Architecture Search}
We then use the Chameleon methodology \cite{Dai2019} to find an optimal design given an architecture search space. 
Briefly, the Chameleon algorithm iteratively samples points from the search space to train an accuracy predictor. This accuracy predictor is used to accelerate a genetic search to find a model that maximizes predicted accuracy while satisfying specified resource constraints. 
In this setting, we used a search space which varies the channel expansion factor and number of output channels per block resulting in $3.4 \times 10^{22}$ possible architectures. 
We set a FLOP constraint on the model architecture and can vary this constraint in order to achieve different operating points. 
The total time to search was approximately three days using 800 Tesla V100 GPUs.

\paragraph{Quantization}
The result of the architecture search is an optimized model with a reduced FLOP count and a lower number of parameters. 
As our model is friendly for low-bit precision computation, we further improve the model by
quantizing the 32-bit floating point parameters and activations \cite{Choi2018Pact} to 8-bit integers. 
This achieves a $4\times$ model size reduction as well as a reduction in inference latency and has been shown to result in only a small accuracy loss in other tasks \cite{Dai2019}. 
We use a standard linear quantizer on both the model parameters and the activations. 
Furthermore, we utilize Quantization-Aware Training (QAT) in order to determine the quantization parameters \cite{Jacob2018} so that performance translates between training and inference.
Our architecture is particularly amenable to quantization and QAT, because it only contains standard 1x1 convolution, depth-wise convolution, BatchNorm, ReLU and resize operations and both convolutions are memory-bounded. 
The model can be further simplified by fusing BatchNorm operations with convolutions and ReLU can be handled by fixing the lower bound of the quantization parameters to zero. 
The convolution and resize operators are the only operators retained in the final model.

\paragraph{Training Details}

We train the network with the MegaDepth dataset, and the scale-invariant data loss and the multi-scale scale-invariant gradient loss proposed by Li and Snavely ~\shortcite{Li2018}, but exclude the ordinal loss term.
The training runs for 100 epochs using minibatches of size 32 and the Adam optimizer with $\beta_1 = 0.5$ and $\beta_2 = 0.999$.
The ground truth depth maps in the MegaDepth dataset do not include depth measurements for the sky region.
We found that using this data as-is led to a network that would not reliably place the sky in the background.
To overcome this limitation we leverage PSPNet \cite{Zhao2017} to identify the sky region in the images.
We then replace any missing depth information in the sky region with twice the maximal depth observed in the depth map.
Intuitively, this forces the sky to have the largest depth in the scene for all MegaDepth images.

To prevent overfitting to specific camera characteristics in our data we perform data augmentation by varying color saturation, contrast, image brightness, hue, image area, field of view, and left-right flipping.
Specifically, all images have an aspect ratio of 4:3 (or its inverse) and are first uniformly resized so the short side is $288\alpha$ pixels long.
$\alpha$ is a uniform random sample in $\left[1, 1.5\right]$ to avoid overfitting to the field of view of the training data.
The resize operation uses nearest neighbor point sampling for both the depth map and the image (it, interestingly, performed better than proper anti-aliased sampling).
Next, we select a random crop of size $(288, 288)$ from the resized image and depth map. Finally, we apply random horizontal flipping to the $(288, 288)$ image and depth map crops.

Another deficiency we found with early networks was that they failed to generalize to images from lower quality cameras.
The images in the MegaDepth dataset are well-exposed, but we often found that other cameras fail to provide such good exposure.
To enable robustness to image color variations, we combine the above data augmentation with the following color data augmentation.
We vary the brightness of the image by a gamma adjustment with a uniformly distributed factor in $[0.6, 1.4]$.
Similarly, we vary the contrast uniformly between 60\% and 100\% \rev{(i.e., blending with a middle-gray image)} as well as the color saturation of the image.
Finally, images are converted to the HSV colorspace and the hue is rotated by a uniformly distributed value in $[-25^{\circ}, 25^{\circ}]$ before being converted back to RGB.

\ignorethis{
Architecture Search \cite{Wu2018},
minimizing this equation.
\begin{equation}
\min_{a \in \mathcal{A}} \min_{w_a} \mathcal{L}\!\left( a, w_a \right)	
\end{equation}

\johannes{Describe how AS works}

we use state-of-the-art machine learning techniques to predict a plausible depth map from only a single color image as input.

 accuracy improvement comes at the cost of
higher computational complexity, making it more challenging to deploy ConvNets to mobile devices, on which computing capacity is limited. Instead of solely focusing on
accuracy, recent work also aims to optimize for efficiency,
especially latency. 

Quantization

\paragraph{Design space $\mathcal{A}$}

In this work, we construct a layer-wise search space with
a fixed macro-architecture, and each layer can choose a different block. The macro-architecture is described in Table
1. The macro architecture defines the number of layers and
the input/output dimensions of each layer.

for each layer of the network, we can choose a different kernel size from \{1, 3, 5\} and a different filter number from \{32, 64, 128, 256, 512\}. )

\paragraph{Loss function $\mathcal{L}$}

\paragraph{Search algorithm}

Describe training procedure (epochs, duration, what machines)

\paragraph{Data augmentation}

To prevent overfitting to iPhone camera characteristics in our data we perform data augmentation by varying color saturation, contrast, image brightness, image area, field of view, and left-right flipping.
Specifically, all images have an aspect ratio of 4:3 (or inverse) and are first uniformly resized so the short side is $256\alpha$ pixels long.
$\alpha$ is a uniform random sample in $\left[1, 2\right]$ to avoid overfitting to the field of view of the training data.
The resize operation uses nearest neighbor point sampling for both the depth map and the image (it, interestingly, performed better than proper anti-aliased sampling).
Next, we select a random crop of size $(240, 240)$ from the resized image and depth map. Finally, we apply random horizontal flipping to the $(240, 240)$ image and depth map crops. 

Another deficiency we found with early networks was that they failed to generalize to images from lower quality cameras.  Our Portrait mode photo dataset is very well-exposed, but we often found that other cameras fail to provide such good exposure.  To enable robustness to image color variations, we combine the above data augmentation with the following color data augmentation. We vary the brightness of the image by a uniformly distributed factor of $\gamma \sim \mathcal{U}(0.6, 1.4)$. Similarly, we vary the contrast uniformly between 60\% and 100\% as well as the color saturation of the image.  

\paragraph{Result}

Describe optimized architecture (parameters, etc.)
}

\ignorethis{
There are an order of magnitude more phones with regular single-lens cameras than those with dual-lens or structured light depth cameras.
In order to expand the reach of our system to all those devices, we use state-of-the-art machine learning techniques to predict a plausible depth map from only a single color image as input.
This process, called monocular depth estimation, is easily integrated into our system and run before the main 3D photo generation pipeline (Section~\ref{sec:dual_lens}).
This approach does not only enable 3D photo generation with any kind of phone camera, it also opens up the possibility of converting legacy content to 3D such as historical photos as shown in the supplemental video.

The MegaDepth system by Li and Snavely ~\shortcite{Li2018} is the state-of-the-art approach for monocular depth estimation.
It uses a convolutional neural network (CNN) and large quantities of internet photos for training data, and provides substantially improved generalization capability compared to previous work.
However, there are, in particular, two limitations that are problematic for 3D photo generation:
(1) depth discontinuities are often poorly-defined due to excessive blurring, and
(2) often isolated whole body parts have significant erroneous depth, e.g., disconnecting a person's head from their torso at the collar of their shirt.
The latter limitation likely results from the fact that the MegaDepth training dataset does not contain accurate depth data for people.
This is because it uses multiple images to triangulate 3D structures for depth, and people are often not visible in precisely the same pose between photos in unstructured Internet photo collections.
We alleviate these limitations by using an additional new dataset created from publicly shared 3D photos.

We leverage the hourglass CNN proposed by Chen et~al.~\shortcite{Chen2016} and train this network using the combination of two datasets: (1) The MegaDepth dataset introduced by Li and Snavely ~\shortcite{Li2018} and (2) a Portrait mode dataset of 250,000 iOS Portrait mode photos with embedded depth maps, uploaded by users of a social networking website for public consumption.  Briefly, the MegaDepth dataset contains two kinds of data annotations, some with per-pixel depth and others with only foreground-background ordinal constraints.  In particular, permanent structures such as buildings have per-pixel depth available whereas transient objects such as people only have ordinal constraints available.  Since our Portrait mode dataset contains per-pixel depth everywhere, including transient objects, the ordinal examples in MegaDepth no longer provide additional value.  Therefore, we train the hourglass CNN leveraging the scale-invariant data loss and the multi-scale scale-invariant gradient loss proposed by Li and Snavely ~\shortcite{Li2018}, but exclude the ordinal loss term.  In total, we use 380,000 photos for training.

We train the hourglass network for 20 epochs using minibatches of size 32 and the Adam optimizer with $\beta_1 = 0.5$ and $\beta_2 = 0.999$. The ground truth depth maps in the MegaDepth do not include depth measurements for the sky region. We found that using this data as-is led to a network that would not reliably place the sky in the background.  To overcome this limitation we leverage PSPNet ~\cite{Zhao2017} to identify the sky region in the images. We then replace any missing depth information in the sky region with twice the maximal depth observed in the depth map. Intuitively, this forces the sky to have the largest depth in the scene for all MegaDepth images. No additional processing was necessary to make use of the depth in the Portrait mode dataset.

To prevent overfitting to iPhone camera characteristics in our data we perform data augmentation by varying color saturation, contrast, image brightness, image area, field of view, and left-right flipping.
Specifically, all images have an aspect ratio of 4:3 (or inverse) and are first uniformly resized so the short side is $256\alpha$ pixels long.
$\alpha$ is a uniform random sample in $\left[1, 2\right]$ to avoid overfitting to the field of view of the training data.
The resize operation uses nearest neighbor point sampling for both the depth map and the image (it, interestingly, performed better than proper anti-aliased sampling).
Next, we select a random crop of size $(240, 240)$ from the resized image and depth map. Finally, we apply random horizontal flipping to the $(240, 240)$ image and depth map crops. 

Another deficiency we found with early networks was that they failed to generalize to images from lower quality cameras.  Our Portrait mode photo dataset is very well-exposed, but we often found that other cameras fail to provide such good exposure.  To enable robustness to image color variations, we combine the above data augmentation with the following color data augmentation. We vary the brightness of the image by a uniformly distributed factor of $\gamma \sim \mathcal{U}(0.6, 1.4)$. Similarly, we vary the contrast uniformly between 60\% and 100\% as well as the color saturation of the image.  

\begin{table}[t]
\begin{tabular}{lcc}
 &  \textbf{MegaDepth} &  \textbf{Our network}\\
 \hline
mean si-RMSE & 0.4954
& 0.4488 \\
median si-RMSE & 0.3485 
& 0.3032 \\
 Silhouette & 0.2271 
 & 0.1261 
 \\
 Disparity &  0.0328 
 & 0.0129 
 \\
\end{tabular}
\caption{Comparison of our CNN for 3D photo generation with the state of the art MegaDepth depth CNN~\cite{Li2018}. See text for a description of the ``silhouette'' and ``disparity'' metrics.
\label{tab:evalMD}}
\end{table}

We created a separate dataset of 679 dual camera images to evaluate the quality of our trained network with respect to representative 3D photos. To measure the impact of our modified loss function and the augmented training dataset, we evaluated the network using the scale-invariant RMSE (si=RMSE) of Li and Snavely~\shortcite{Li2018}.

In 3D photos, the exact depth values are often less salient than the depth ordering and placement of discontinuities.
We evaluate two additional metrics that capture these specifcic characteristics that are important for our application:

\begin{description}
\item{Silhouette:}
we run the depth pre-processing stage of our pipeline (Section~\ref{sec:preproc}) and evaluate the fraction of mis-matched silhouette pixels (allowing slight shifts of 5 pixels when match ground truth vs.~estimated depth silhouettes).

\item{Disparity:}
We evaluate the relative depth variation at a local level by comparing normalized blocks of depth values.
We consider overlapping 48 x 48 pixel blocks with a stride of 16, normalize their depth by subtracting the mean, and evaluate the absolute depth differences.
The final measure reports the average difference of all blocks.
\end{description}}

%% file: figures/tiefenrausch_scheme.tex
\begin{figure}
\centering%
\rev{\includegraphics[width=\linewidth]{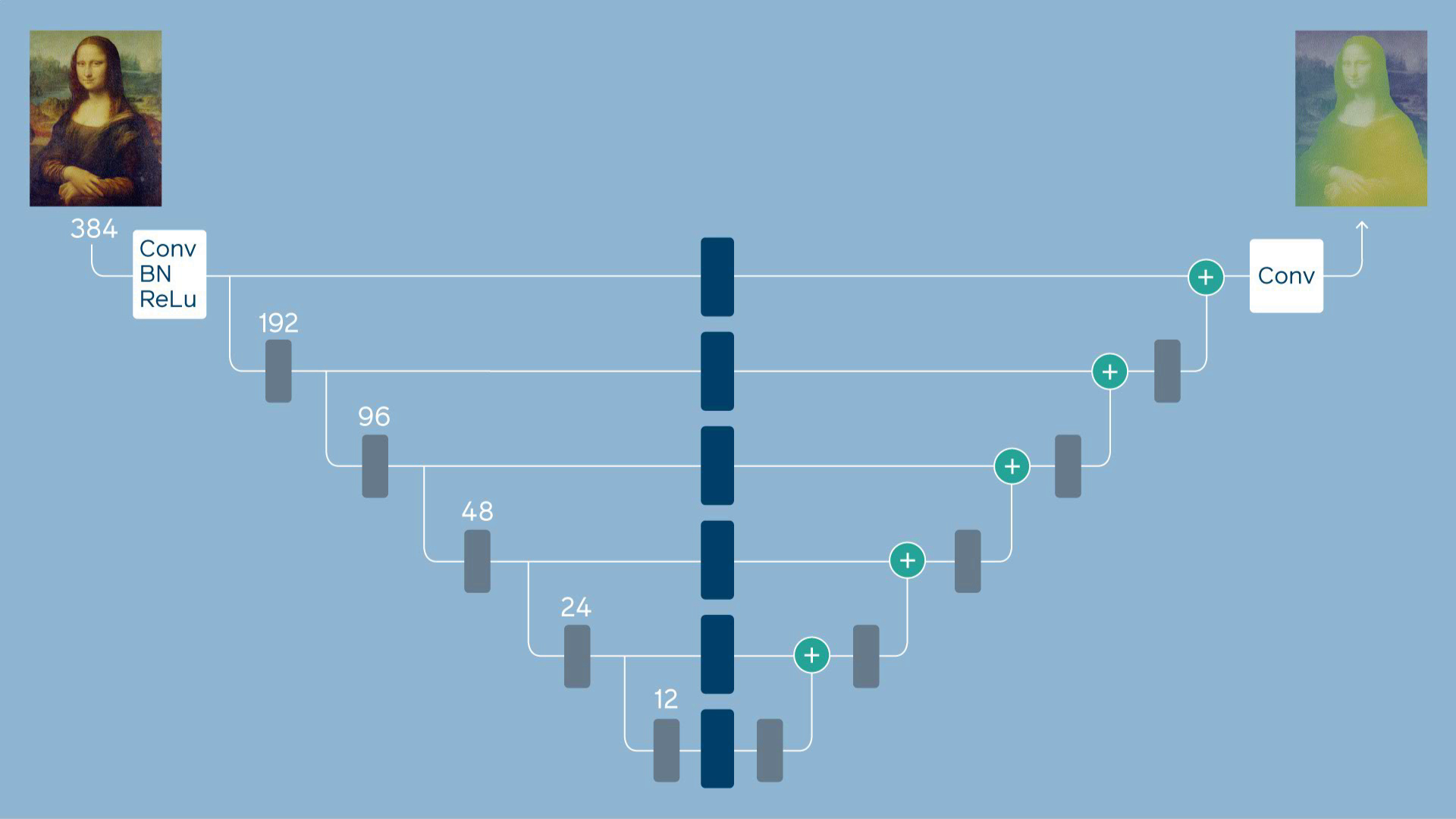}}\\
\caption{Depth estimation network schematic. \rev{Gray TR blocks are used in down-/up-sampling passes and blue TR blocks are used to preserve spatial resolution.} TR Blocks are defined in Fig. \ref{fig:block_structure}.
Mona Lisa, by Leonardo da Vinci (public domain).}
\label{fig:tr_scheme}
\end{figure}

%% file: figures/block_structure.tex
\begin{figure}
  \centering%
        \includegraphics[keepaspectratio,height=6cm]{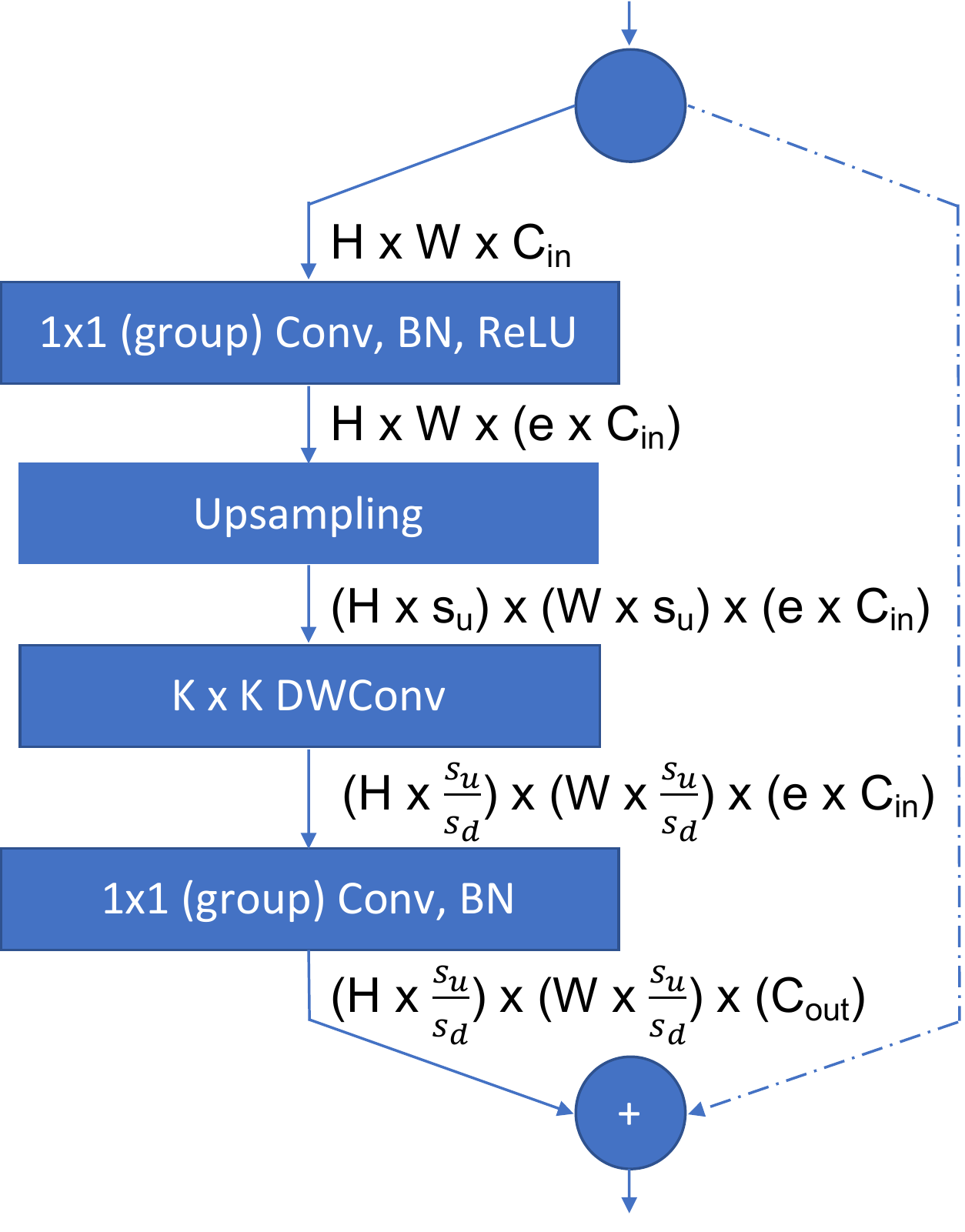}
	\def\desc{Block structure used to create the depth estimation architecture.}
	\caption{
		\desc~
		\rev{$s_u$ and $s_d$ refer to the up and down sampling scale factors, respectively, and $e$ is the channel expansion factor.
		Refer to the text for details.}
		}
	\Description[\desc]{\desc}
	\undef\desc
  \label{fig:block_structure}
\end{figure}

%% file: tex/layering.tex
\subsection{Lifting to Layered Depth Image}
\label{sec:layering}

\input{figures/depth_filtering.tex}

Now that we have obtained a dense depth map, we are ready to lift the image to the LDI representation.
This will allow us to express multiple layers, so we can show detail in parallax regions.
These are details that have not been observed in the input view, they have to be synthesized.

After discussing depth pre-processing in Section~\ref{sec:preproc}, we will discuss hallucinating new geometry in parallax regions in Section~\ref{sec:depth_hallucination}, and finally inpainting of color on the new geometry in Section~\ref{sec:inpainting}.



\subsubsection{Depth Pre-processing}
\label{sec:preproc}

The most salient geometric feature in 3D photos are depth discontinuities.
At those locations we need to extend and hallucinate new geometry behind the first visible surface (as will be explained in the next section).
The depth images obtained in the previous section are typically over-smoothed due to the regularization inherent to the machine learning algorithms that produced them.
This smoothing ``washes out'' depth discontinuities over multiple pixels and often exhibit spurious features that would be difficult to represent (Figure~\ref{fig:depth_filtering}b).
The goal of the first algorithm stage is to de-clutter depth discontinuities and sharpen them into precise step edges.

We first apply a \emph{weighted} median filter\footnote{\rev{i.e., sort the samples by value and find the one whose sums of preceding weights and following weights are closest to being equal.}} with a $5 \times 5$ kernel size.
Depth values within the kernel are Gaussian-weighted by their disparity difference to the center pixel (using $\sigma_\textit{\!disparity}=0.2$).
The weighting of the filter is important to preserve the localization of discontinuities, and, for example,  avoid rounding off corners.
Since we are interested in forcing a decision between foreground and background, we disable the weights of pixels near the edge (i.e., pixels that have a neighbor with more than $\td=0.05$ disparity difference.)

This algorithm succeeds in sharpening the discontinuities.
However, it occasionally produces isolated features at middle-depth values (Figure~\ref{fig:depth_filtering}c).
We perform a connected component analysis (with threshold $\td$) and merge small components with fewer than 20 pixels into either foreground or background, whichever has a larger contact surface (Figures~\ref{fig:depth_filtering}d).


\subsubsection{Occluded Surface Hallucination}
\label{sec:depth_hallucination}

\input{figures/depth_expand.tex}
\input{figures/atlas.tex}

The goal of this stage is to ``hallucinate'' new geometry in occluded parts of the scene.
We start by lifting the depth image onto an LDI to represent multiple layers of the scene.
Initially, the LDI has a single layer everywhere and all pixels are fully connected to their neighbors, except across discontinuities with a disparity difference of more than $\td$.

To create geometry representing occluded surfaces, 
we next extend the geometry on the \emph{backside} of discontinuities iteratively \emph{behind} the front-side by creating new LDI pixels.
A similar algorithm has been employed by Hedman and Kopf~\shortcite{Hedman2018}.
However, their algorithm has an important limitation:
pixels are allowed to extend in all directions (as long as they remain hidden behind the front layer).
This causes frequent artifacts at T-junctions, i.e., where background, midground, and foreground meet:
the midground grows unrestrained, expanding the foreground discontinuity and creating a cluttered result (Figure~\ref{fig:depth_expand}b).
The authors reduce the undesired excess geometry by removing all but the nearest and farthest layers anywhere in the LDI.
However, this creates disconnected surfaces (Figure~\ref{fig:depth_expand}c).
We resolve these problems by grouping discontinuities into \emph{curve-like} features and inferring spatial constraints to better shape their growth (Figure~\ref{fig:depth_expand}d).
We group neighboring discontinuity pixels together, but not across junctions (see color coding in Figure~\ref{fig:depth_expand}a).
At this point, we remove spurious (shorter than 20 pixels) groups from consideration.

In one extension iteration, each group grows together as one unit, creating a one pixel wide ``wave front'' of new LDI pixels.
To avoid the previously mentioned cluttering problem, we restrain curves from growing beyond the perpendicular straight line at their end points (dotted lines in Figure~\ref{fig:depth_expand}a).
3-way intersections deserve special consideration: at these points there are 3 different depths coming together, but we are \emph{only} interested in constraining the \emph{mid}ground, while the \emph{back}ground should be allowed to freely grow under both of the other layers.
Therefore, we only keep the one of the three constraints at 3-way intersections that is associated with the mid-/foreground discontinuity (Figure~\ref{fig:depth_expand}a).

The depth of newly formed pixels is assigned an average of their neighbors, and the color is left undefined for now (to be inpainted in the next section).
Intersecting groups are merged if their disparity difference is below $\td$.
We run this expansion algorithm for 50 iterations to obtain a multi-layered LDI with sufficient overlap for displaying it with parallax.

%% file: figures/depth_filtering.tex
\newlength\fleb
\setlength\fleb{5.9cm}
\newlength\flec
\setlength\flec{2.34cm}
\newlength\fled
\setlength\fled{6.5cm}
\newlength\flee
\setlength\flee{1.8cm}

\begin{figure}
  \centering%
%
%
\ignorethis{	\parbox[t]{\fleb}{\vspace{0mm}\centering%
		\includegraphics[width=\fleb,trim=0 150 0 110,clip]%
				{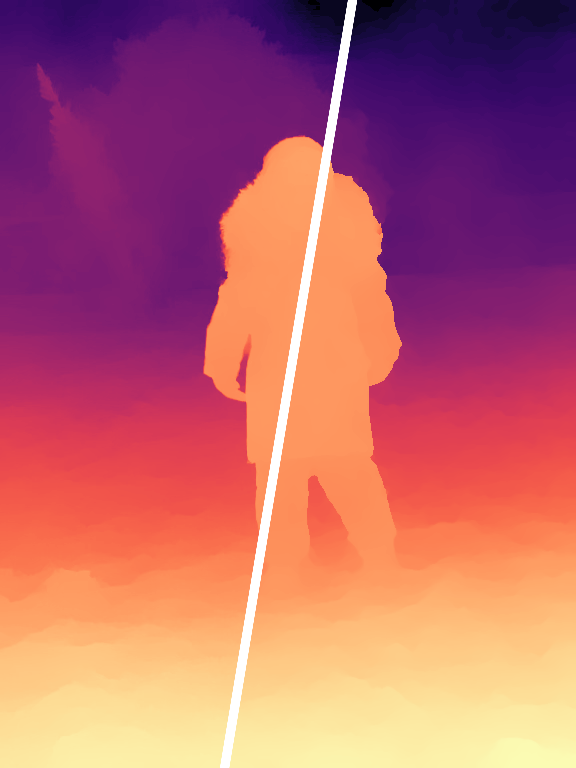}\vspace{-1mm}\\%
		(a) Raw input / processed depth image%
	}%
	\hfill%
	\parbox[t]{\flec}{\vspace{0mm}\centering%
		\fbox{\includegraphics[width=\flec]%
				{figures/depth_filtering/raw.png}}\vspace{-1mm}\\%
		(b) Raw input\vspace{1mm}\\%
		\fbox{\includegraphics[width=\flec]%
				{figures/depth_filtering/proc.png}}\vspace{-1mm}\\%
		(c) Processed%
	}}%
%
%
	\parbox[t]{\fled}{\vspace{0mm}\centering%
		\includegraphics[width=\fled,trim=0 130 0 70,clip]%
				{figures/depth_filtering/depth_slash.png}\vspace{-1mm}\\%
		(a) Raw / cleaned depth image%
	}%
	\hfill%
	\parbox[t]{\flee}{\vspace{0mm}\centering%
		\fbox{\includegraphics[width=\flee]%
				{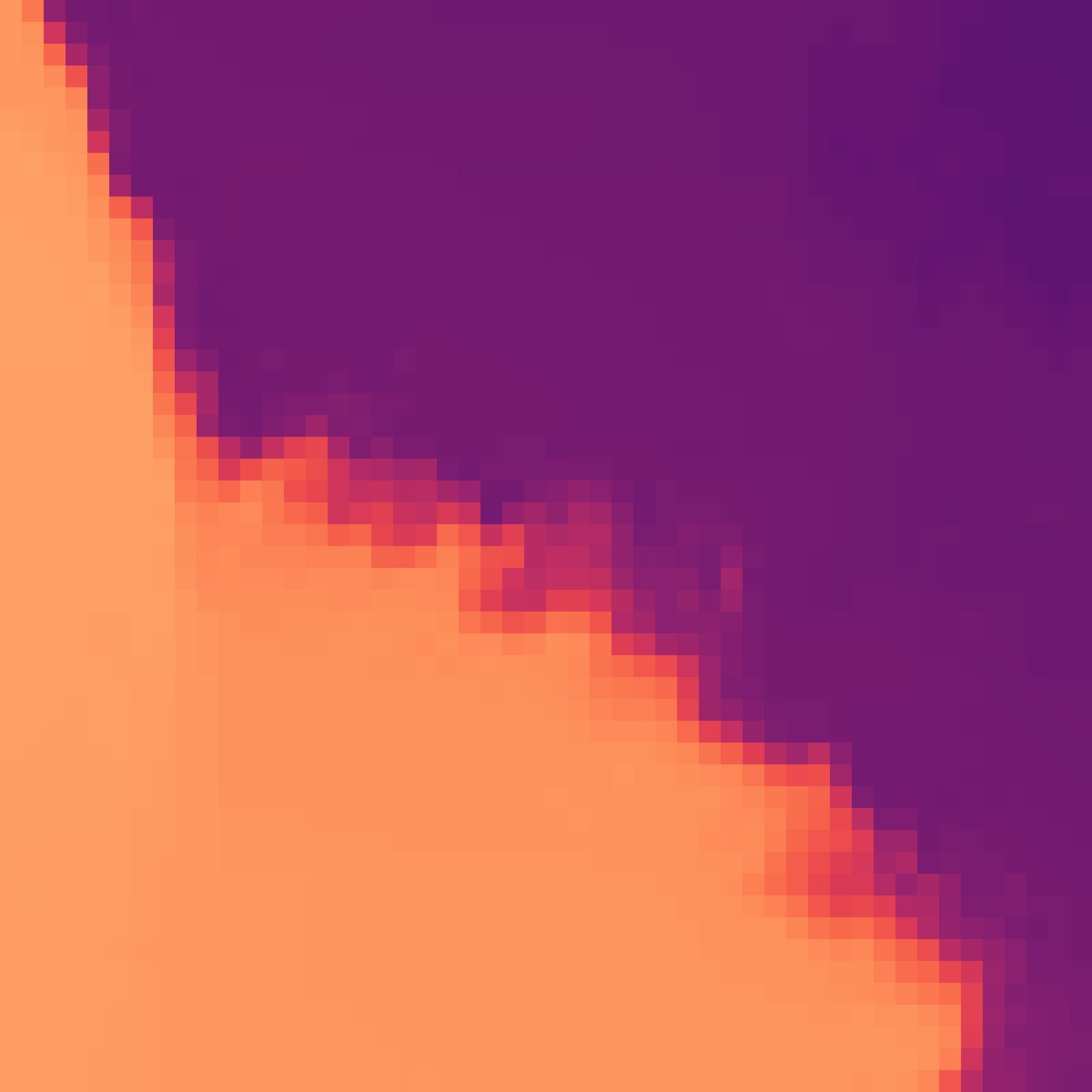}}\vspace{-1mm}\\%
		(b) Raw\vspace{1mm}\\%
		\fbox{\includegraphics[width=\flee]%
				{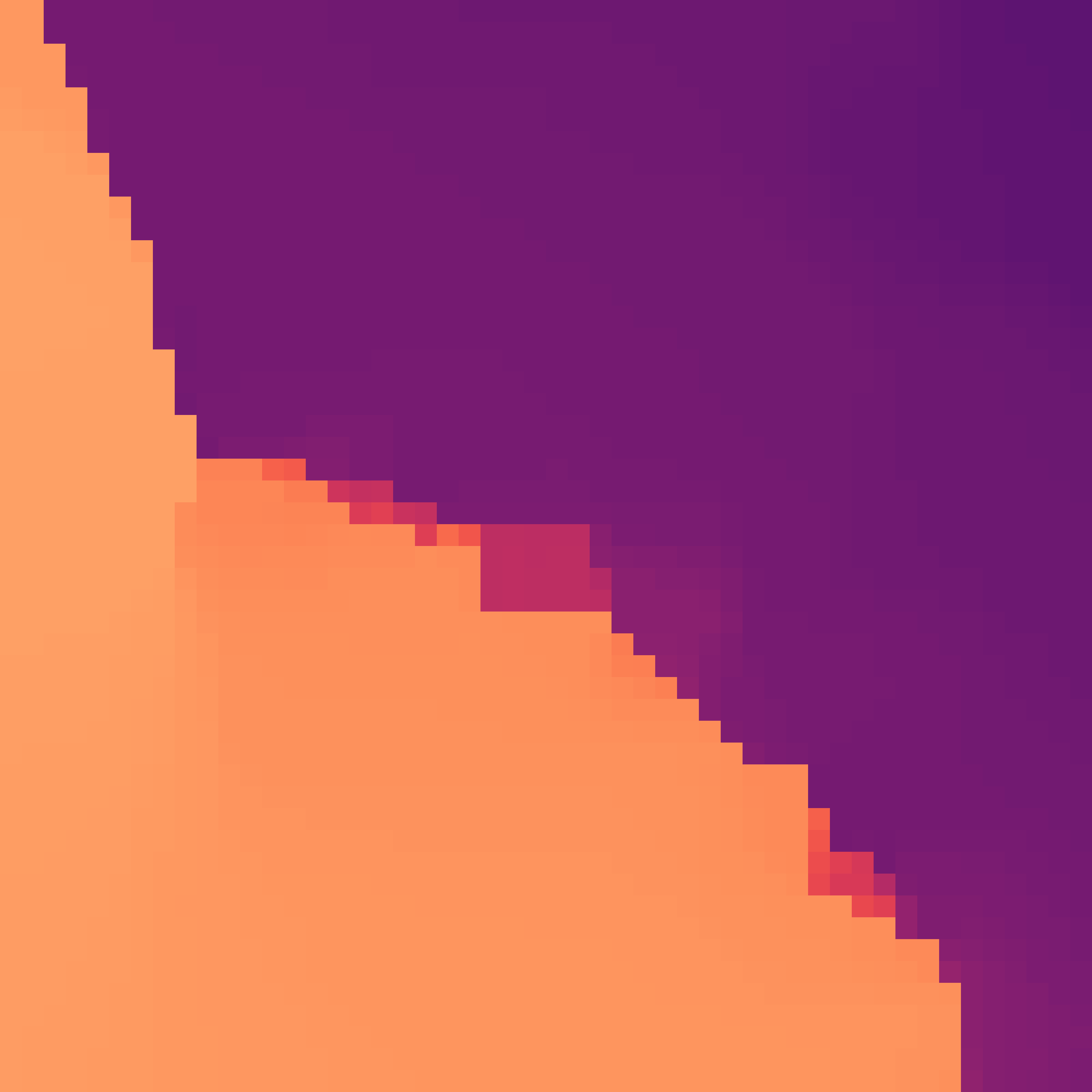}}\vspace{-1mm}\\%
		(c) Filtered\vspace{1mm}\\%
		\fbox{\includegraphics[width=\flee]%
				{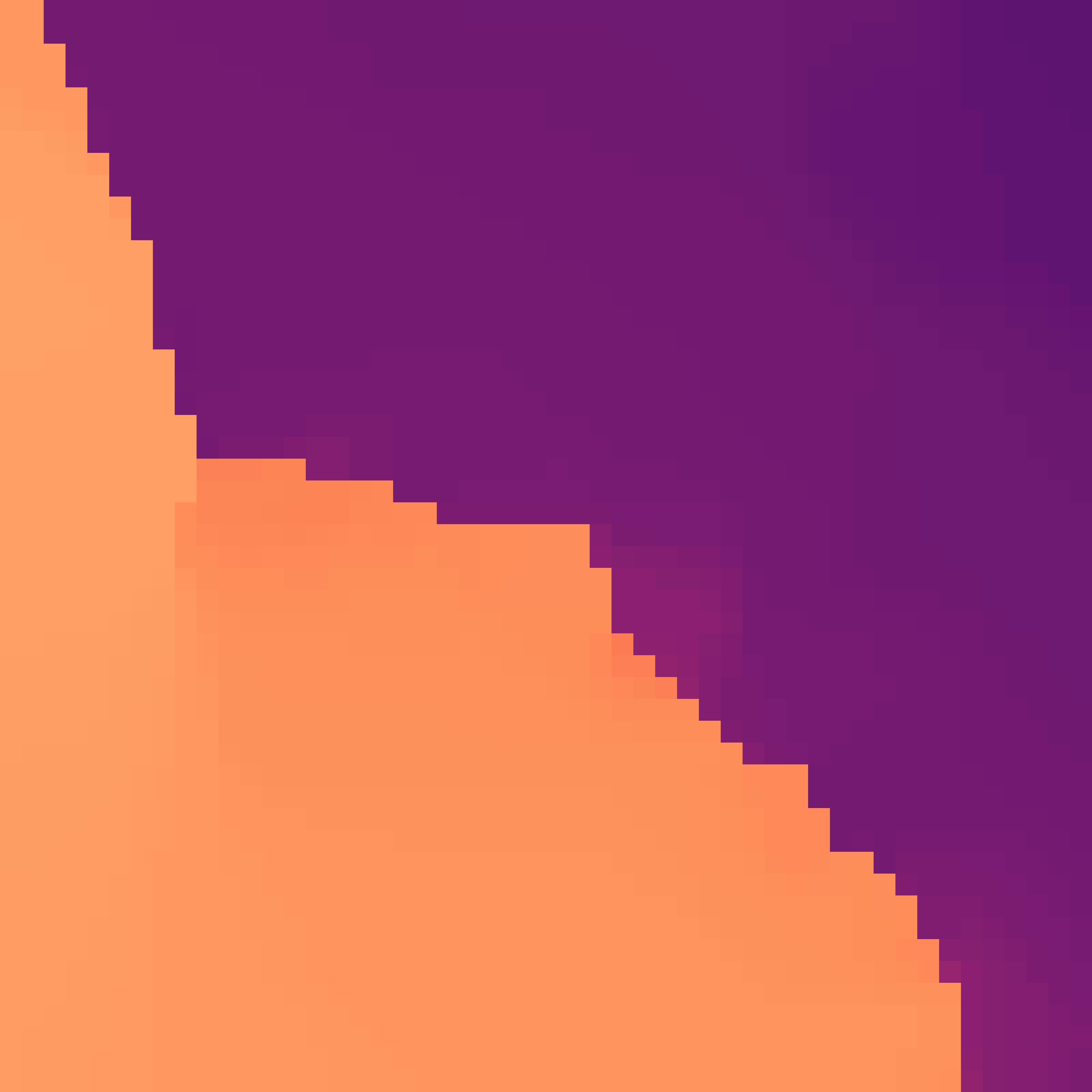}}\vspace{-1mm}\\%
		(d) Cleaned%
	}%
	\vspace{-3mm}\\%
	\def\figcap{
Depth image before and after cleaning (a).
Discontinuities are initially smoothed out over multiple pixels.
Weighted median filter sharpens them successfully in most places (c).
We fix remaining isolated features at middle-values using connected component analysis (d).}
	\caption{\figcap}
	\Description[Depth image before and after cleaning.]{\figcap}
	\undef\figcap
  \label{fig:depth_filtering}
\end{figure}

%% file: figures/depth_expand.tex
\ignorethis{
\newlength\fmda
\setlength\fmda{5.5cm}
\newlength\fmdb
\setlength\fmdb{2.7cm}

\begin{figure}
  \centering%
	\parbox[t]{\fmda}{\vspace{0mm}\centering%
		\includegraphics[width=\fmda,trim=0 150 200 0,clip]%
				{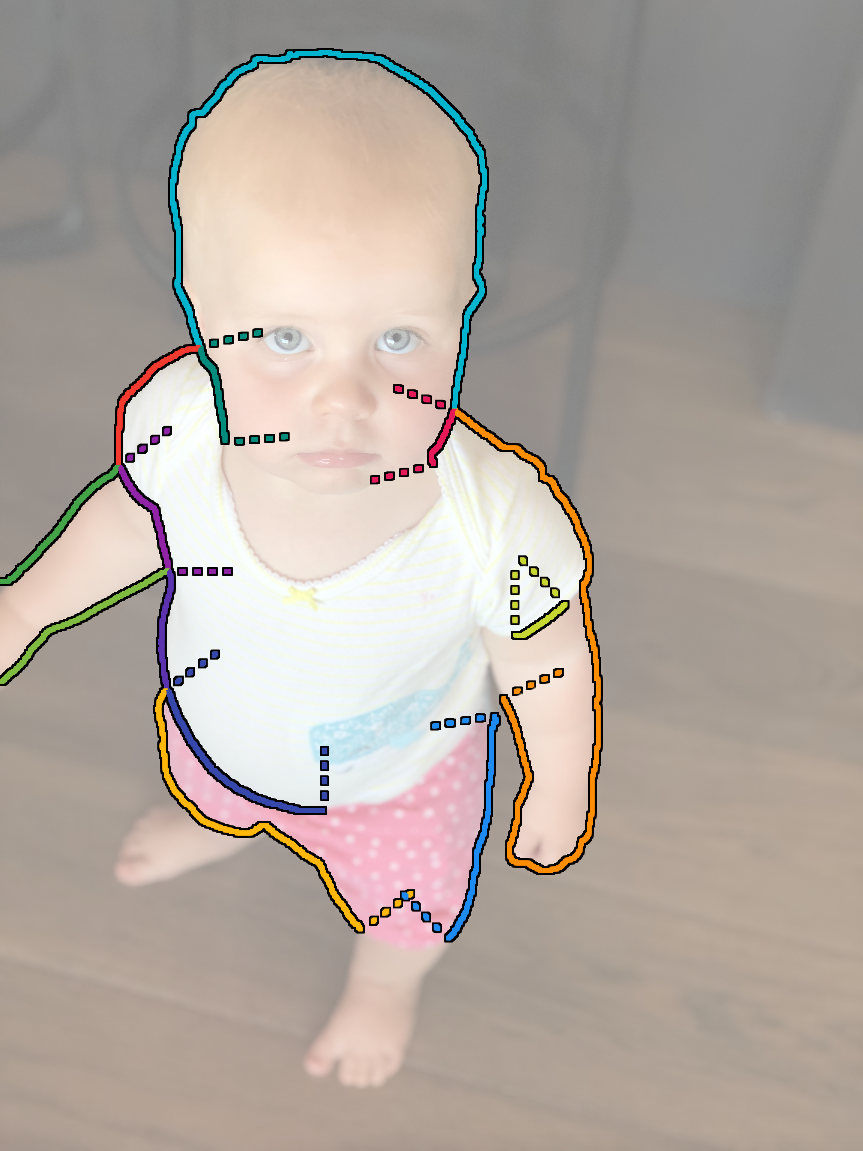}\vspace{-1mm}\\%
		Expansion constraints%
	}%
	\hfill%
	\parbox[t]{\fmdb}{\vspace{0mm}\centering%
		\fbox{\includegraphics[width=\fmdb]%
				{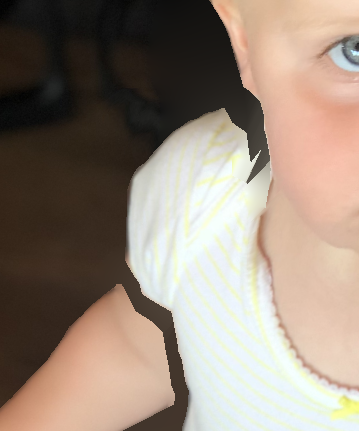}}\vspace{-1mm}\\%
		Without constraints\vspace{1mm}\\%
		\fbox{\includegraphics[width=\fmdb]%
				{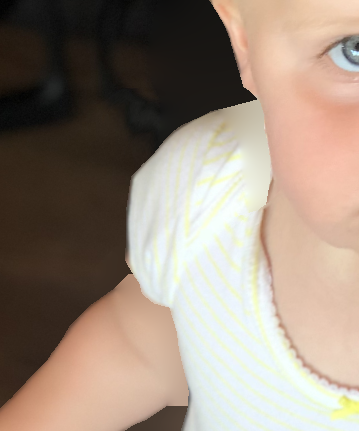}}\vspace{-1mm}\\%
		With constraints%
	}%
\ignorethis{  \jsubfig{\includegraphics[height=\fmda]{figures/depth_expand/constraints.png}}{(a) Fine outline}%
	\hfill%
  \jsubfig{
	  \includegraphics[width=\fmdb]{figures/depth_expand/cracked_crop.png}\\%
		\includegraphics[width=\fmdb]{figures/depth_expand/fixed_crop.png}%
	}{(a) Fine outline}}%
	\vspace{-3mm}\\%
	\caption{Depth expansion.}
  \label{fig:depth_expand}
\end{figure}}

%
%
\newlength\fmda
\setlength\fmda{6.0cm}
\newlength\fmdb
\setlength\fmdb{2.4cm}

\begin{figure}
  \centering%
	\parbox[t]{\fmda}{\vspace{0mm}\centering%
		\includegraphics[width=\fmda,trim=0 190 200 30,clip]%
				{figures/depth_expand/constraints.png}\vspace{-1mm}\\%
		\small (a) Expansion constraints%
	}%
	\hfill%
	\parbox[t]{\fmdb}{\vspace{0mm}\centering%
		\fbox{\includegraphics[width=\fmdb,trim=0 10 0 48,clip]%
				{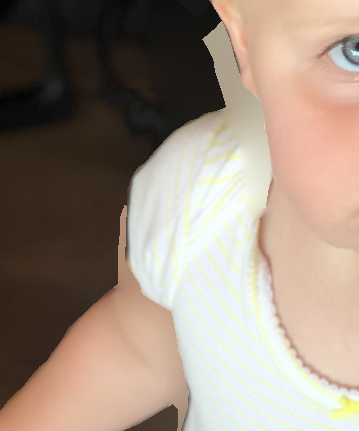}}\vspace{-1mm}\\%
		\small (b) Instant3D \emph{N}-layer\vspace{1mm}\\%
		\fbox{\includegraphics[width=\fmdb,trim=0 10 0 48,clip]%
				{figures/depth_expand/cracked_crop.png}}\vspace{-1mm}\\%
		\small (c) Instant3D \emph{2}-layer\vspace{1mm}\\%
		\fbox{\includegraphics[width=\fmdb,trim=0 10 0 48,clip]%
				{figures/depth_expand/fixed_crop.png}}\vspace{-1mm}\\%
		\small (d) Our method%
	}%
	\vspace{-3mm}\\%
	\def\figcap{
Expanding geometry on the back-side of discontinuities into occluded parts of the scene.
Previous work \cite{Hedman2018} produces artifacts at T-junctions: either extraneous geometry if left unconstrained (b) or cracked surfaces when using their suggested fix (c).
We improve this by grouping discontinuities into curve-like features (color-coded), and inferring spatial constraints to better shape their growth (dashed lines).
}
  \caption{\figcap}
	\Description[Expanding geometry on the back-side of discontinuities into occluded parts of the scene.]{\figcap}
  \label{fig:depth_expand}
\end{figure}

%% file: figures/atlas.tex
\newlength\fda
\setlength\fda{4.2cm}

\begin{figure*}
\centering%
\jsubfig{%
	\includegraphics[height=\fda]{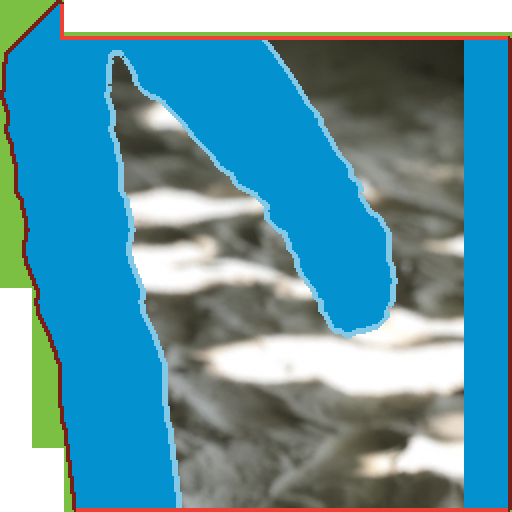}%
	\hspace{1mm}%
	\includegraphics[height=\fda]{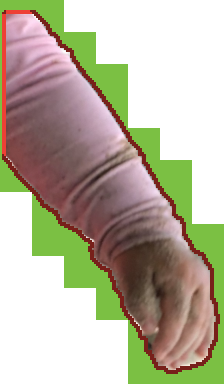}%
 }{(a) Pixel labels}
\hfill%
\jsubfig{%
	\includegraphics[height=\fda]{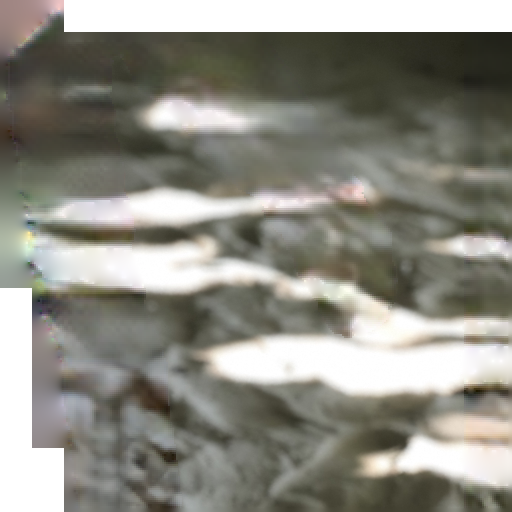}%
	\hspace{1mm}%
	\includegraphics[height=\fda]{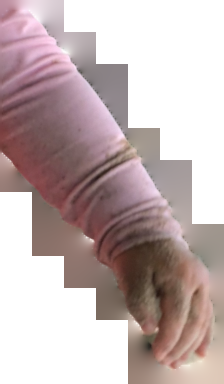}%
 }{(b) Inpainted and padded}%
\hfill%
\jsubfig{%
	\fbox{%
		\includegraphics[height=\fda]{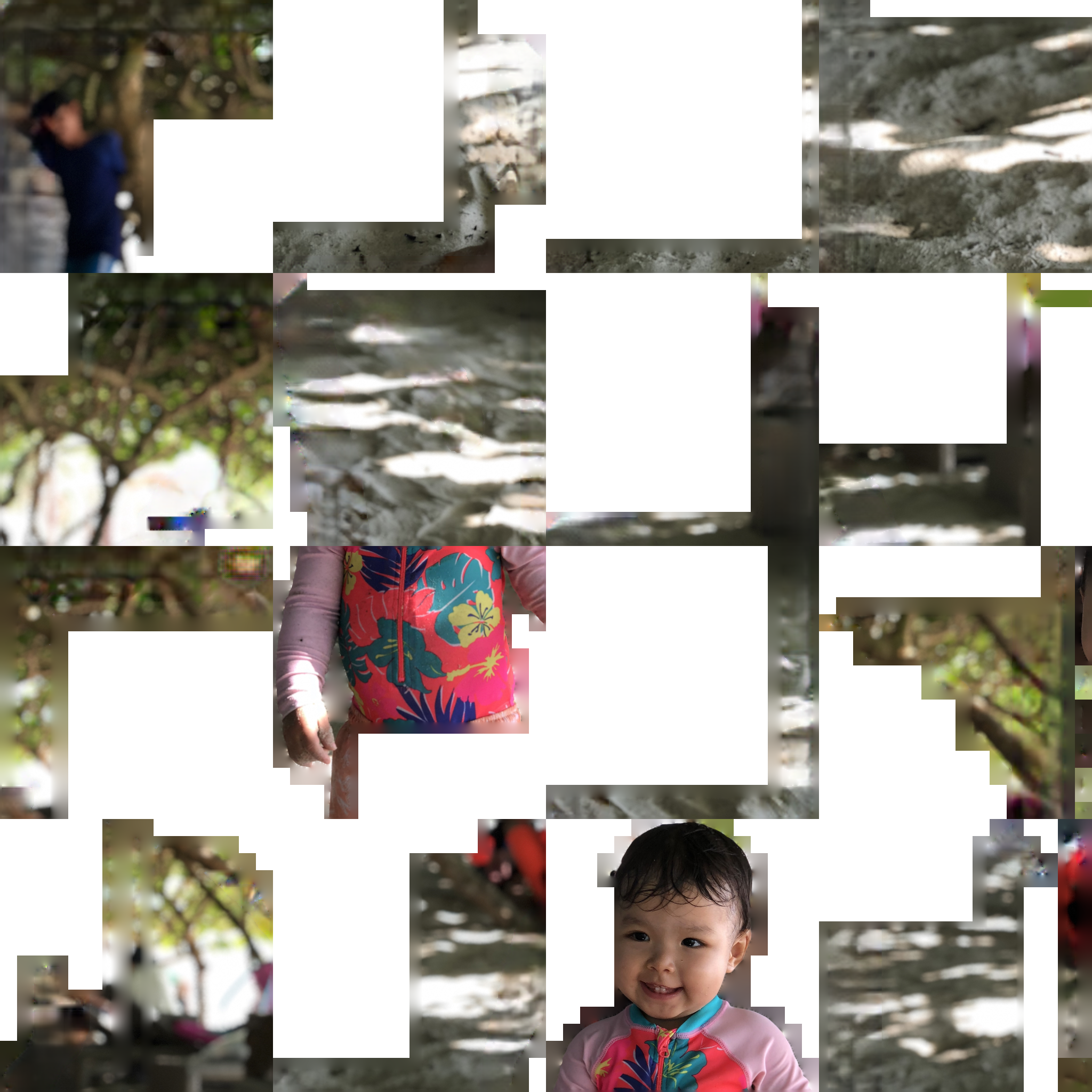}%
	}%
}{(c) Packed texture atlas}%
\vspace{-3mm}\\%
\def\figcap{
Partitioning the layered depth image into charts for texturing.
(a) Pseudo-coloring different kinds of pixels that require inpainting, on two example charts.
Dark blue pixels are occluded, and light blue pixels are on the foreground but close to a discontinuity, and, therefore, contain possibly mixed colors.
Red pixels add padding for texture filtering: dark red pixels (at silhouettes) are inpainted and light red pixels (elsewhere) are copied from adjacent charts.
Green pixels add padding for JPEG macroblocks (see text).
(b) Final inpainted charts.
(c) Packed atlas.
}
\caption{\figcap}
\label{fig:atlas}
\Description[Partitioning the layered depth image into charts for texturing.]{\figcap}
\undef\figcap
\end{figure*}

%% file: tex/inpainting.tex
\subsection{LDI Inpainting}
\label{sec:inpainting}

At this point we have an LDI with multiple layers around depth discontinuities, but it is still missing color values in the parallax regions (i.e., the red pixel in Figure~\ref{fig:teaser}c).
In this section, we discuss the inpainting of plausible colors, so that disocclusions when viewing the 3D photo appear seamless and realistic.

A \naive approach to filling missing regions in disocclusions would be to inpaint them in \emph{screen space}, for example using a state-of-the-art network, such as Partial Conv~\cite{Liu2018ECCV}.
However, this would not lead to desirable results, because
(1) filling each view at runtime would be slow,
(2) the independent synthesis would result in inconsistent views, and, finally,
(3) the result would be continuous on both foreground and background sides of the missing region (while it should only be continuous on the background side), thus leading to strong blur artifacts along the edges.

A better approach would be to inpaint on the LDI structure.
Then, the inpainting could be performed once, each view would be consistent by design, and, since the LDI is explicitly aware of the connectivity of each pixel, the synthesis would be only continuous across truly connected features.
However, one complication is that a LDI does not lend itself easily to processing with a neural network, due to the irregular connectivity structure.
One approach would be, again, to turn to filling projected views and warp the result back onto the LDI.
But this might require multiple iterations from different angles until all missing pixels are covered.

Our solution to this problem uses the insight that the LDI is \emph{locally} structured like a regular image, i.e., LDI pixels are 4-connected in cardinal directions.
By traversing these connections we can aggregate a local neighborhood around a pixel (described below), which allows us to map network operators, such as convolutions, to the LDI.
This mapping, in turn, allows us to train a network \emph{entirely} in 2D and then use the pretrained weights for LDI inpainting, without having done any training with LDIs.

\subsubsection{Mapping the PConv Network to LDI}
\label{sec:pconv_ldi}
We represent the LDI in tensor form as a tuple of a $C \x K$ \texttt{float32} ``value'' tensor $\mathcal{P}$ and a $6 \x K$ \texttt{int32} ``index'' tensor $\mathcal{I}$,  where $C$ is the number of channels, and $K$ the number of LDI pixels.
The value tensor $\mathcal{P}$ stores the colors or activation maps, and the index tensor stores the pixel position $(x, y)$ position and $(\text{left}, \text{right}, \text{top}, \text{bottom})$ neighbor indices for each LDI pixel. We also store a $1 \x K$ binary ``mask'' tensor $\mathcal{M}$ which indicates which pixels are known and which pixels must be inpainted.

The PartialConv~\cite{Liu2018ECCV} network uses a U-Net like architecture~\cite{Ronneberger2015}.
We map this architecture to LDI by replacing every PConv layer with LDIPConv layer which accepts an LDI ($\mathcal{P}, \mathcal{I}$) and mask $\mathcal{M}$ instead of a $C \x H \x W$ color and $1 \x H \x W$ mask image tensor. All value tensors at a level (i.e. scale $1/s$) of the U-Net share the same index tensor $\mathcal{I}_{s}$.
Most operations in the network are point-wise, i.e., the input/output is a single pixel, for example ReLU or BatchNorm; these map trivially to the LDI.
The only non-trivial operations, which aggregate kernels, are 2D convolution (the network uses $3\x3$, $5\x5$, $7\x7$ kernel sizes), down-scaling (convolutions with $\text{stride} = 2$), and up-scaling.

\input{figures/ldi_aggregation.tex}

\paragraph{Convolution:}
We aggregate the 2D convolution kernels by exploring the LDI graph in breadth-first manner:
starting at the center pixel we traverse LDI pixels in up / down / left / right order to greedily fill the kernel elements.
Once a kernel element has been visited we do not traverse to this position again.
If an LDI pixel is unconnected in a direction (e.g., at a silhouette) we treat it as if the pixel was on an image edge i.e. zero-padding. Since we are using partial convolutions and the mask is also zero-padded, this results in partial convolution based padding~\cite{liu2018partialpadding}.
For a $3\x3$ kernel where all LDI pixels are fully connected, the pattern in Figure~\ref{fig:ldi_ops}a emerges.
Figure~\ref{fig:ldi_ops}b shows an example of a $5\x5$ kernel, where some silhouette pixels have no neighbors in certain directions.
In this case, the breadth-first aggregation explores around these ``barriers'', except for the two pixels in the top-right right that cannot be reached in any way and are partial-padded~\cite{liu2018partialpadding}.

\input{figures/ldi_coarsening.tex}

\paragraph{Downscaling or Strided Convolutions:}
In the image-version of the network, downscaling is done by setting a stride of 2 on convolution operations, i.e., for every $2\x2$ block of pixels at the fine scale, only the top-left pixel is retained at the coarser scale.
We implement down-scaling for the LDI in a similar way:
every LDI pixel with $\operatorname{mod}\!\left(x, 2 \right) = \operatorname{mod}\!\left(y, 2\right) = 0$ is retained.
If multiple LDI pixels occupy a $\left(x, y\right)$ position, they will all be retained.
If for two retained pixels there was a length-2 connecting path at the fine scale, they will also be connected a the coarse scale.
Figure~\ref{fig:ldi_coarsening} illustrates this coarsening scheme.

\paragraph{Upscaling:}
In the image-version of the network, upscaling is done with nearest interpolation, i.e., a $2\x2$ block of pixels at the fine scale all take the value of the corresponding 1 pixel at the coarser scale.
We again, emulate this for the LDI: the whole group of LDI pixels that collapsed into a coarse pixel all take its value. We implemented the original PConv network in Caffe2 with the custom convolution and scaling operators.

\subsubsection{Mobile Optimized Inpainting Network}
\label{sec:opt_ldi}

This network enables high-quality inpainting of parallax regions on LDIs.
However, similar to prior work in depth estimation, it is too large and resource intensive for mobile applications. 

In the following, we propose a new architecture, called \emph{Farbrausch} that is optimized in this regard.
We begin with a traditional screen-space (2D) PartialConv network with 5 stages of downsampling.  This network is converted to our LDI representation with our custom operators.  Chameleon Search is used to identify the best set of hyperparameters encoding the number of output channels for each stage of the encoder (and similarly the paired decoder stage).  In particular, FLOP count is traded off against the PartialConv inpainting loss on its validation set~\cite{Liu2018ECCV}.  This hyperparameter search took 3 days on 400 V100 GPUs.  In this time, 150 networks were trained to build the accuracy predictor used in the genetic search.

\ignorethis{
We also added the ability to reduce unnecessary calculations which are not utilized in inpainting hole pixels by utilizing the mask in the convolutional operations.\kevin{Ayush, please expand}}

\ignorethis{
This enables us to apply the network to LDI, but it is large, slow and uses a lot of memory.
(1) the PConv model is large, and too resource intensive for mobile execution;
Modified the architecture to lower the number of parameters, faster, less memory.
Figure~\ref{fig:fr_scheme} shows the modified network architecture.
Since we are irregular image structure, there is another optimization we can do.
PC network computes output for ALL pixels in the 2D image, whether they need inpainting or not.
For FR we can restrict computation to the nodes that we're interested in.
The last layer this is only a small fraction, however, for the other layers it quickly grows.
Still, we save about xx\% computation.

Putting it all together we can now inpaint LDIs.
Figure XX shows an example.}

\begin{figure}
    \centering
    \includegraphics[trim=0 10 0 0,clip,width=\linewidth]{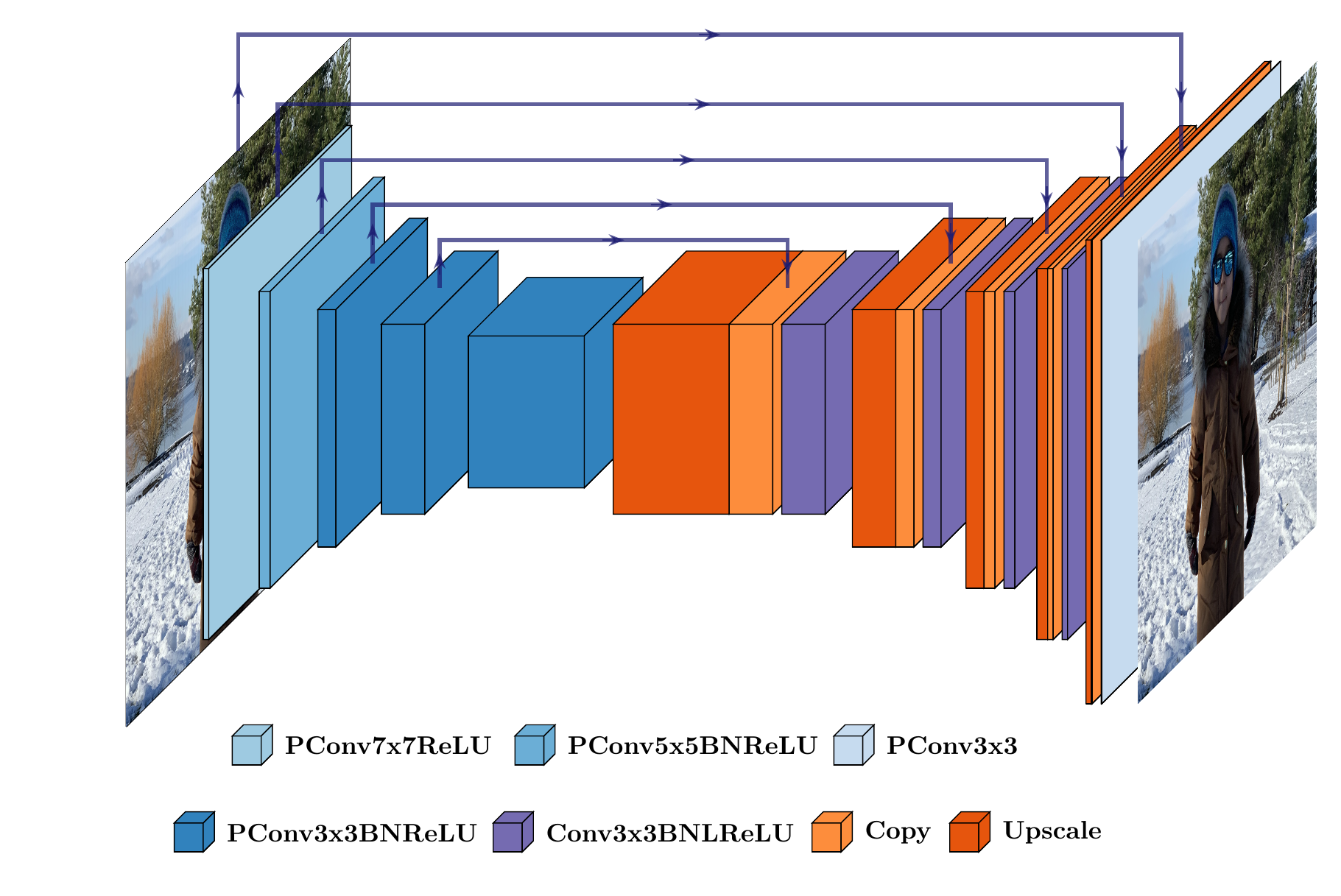}
    \caption{Farbrausch Network}
    \label{fig:my_label}
\end{figure}

%% file: figures/ldi_aggregation.tex
\begin{figure}[t]
\centering%
\jsubfig{~\includegraphics[height=3cm]{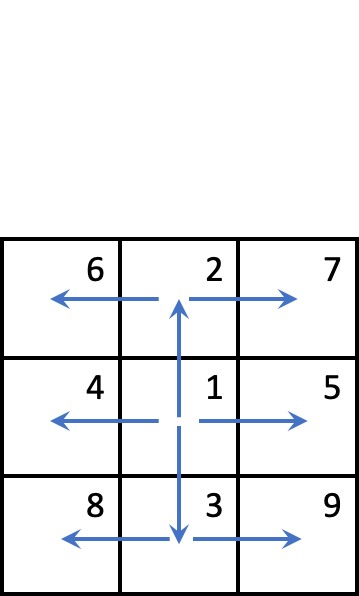}~}{(a) 3x3 kernel\\\small (fully connected)}%
\hspace{0.5cm}%
\jsubfig{\includegraphics[height=3cm]{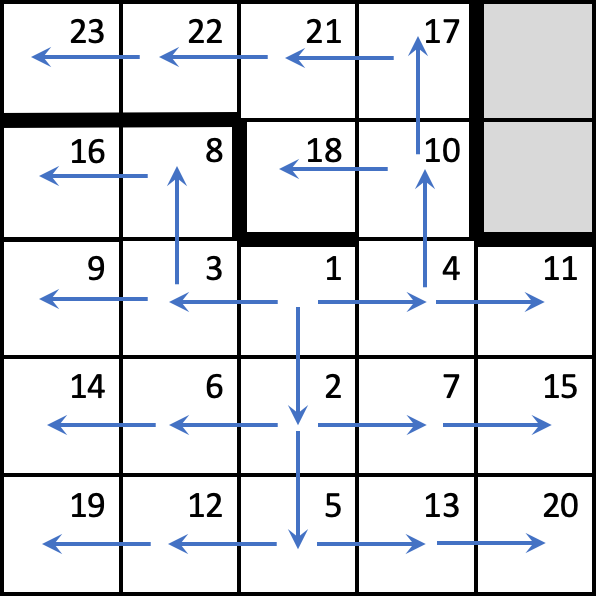}}{(b) 5x5 kernel\\\small (with silhouette pixels)}\\%
\vspace{-3mm}%
\caption{
Aggregating convolution kernels on a LDI with breadth-first exploration.
The numbers indicate the traversal order.
Gray elements cannot be filled and are zero-padded.}
\label{fig:ldi_ops}
\end{figure}

%% file: figures/ldi_coarsening.tex
\begin{figure}[t]
\centering%
\includegraphics[width=\linewidth]{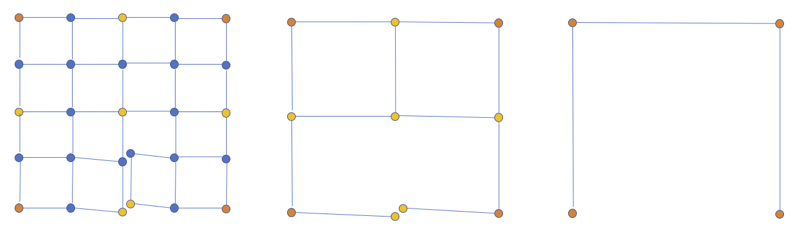}\\
Coarsening\\
\caption{Coarsening scheme for the up-/downscale operators.}
\label{fig:ldi_coarsening}
\end{figure}

%% file: tex/meshing.tex
\input{figures/meshing.tex}

\subsection{Conversion to Final Representation}
\label{sec:meshing}

Now that we have a fully inpainted multi-layer LDI, we are ready to convert it into a textured mesh, which is our final representation.
This is done in two parts: creating the texture (Section~\ref{sec:atlas}), and the mesh generation (Section~\ref{sec:mesh}).


\subsubsection{Texture Atlas Generation}
\label{sec:atlas}

The LDI contains many self-over\-lap\-ping parts and has a complex topology.
Hence, it cannot be mapped to a single contiguous texture image.
We thus partition it into flat \emph{charts} that can be packed into an atlas image for texturing.

\paragraph{Chart generation:}
We use a simple seed-and-grow algorithm to create charts:
the LDI is traversed in scanline order, and whenever a pixel is encountered that is not part of any chart, a new chart is seeded.
We then grow the chart using a breadth-first flood fill algorithm that follows the pixel connections in the LDI, but respects several constraints:
\begin{enumerate}
\item charts cannot fold over in depth, since that would not be representable;
\item we cap the maximum chart size to improve packing efficiency (avoids large non-convex shapes);
\item when encountering pixels at the front side of depth edges (without neighbors in some direction), we mark a range of adjacent pixels across the edge unusable to avoid filtering operations from including pixels from different surfaces. These marked pixels will eventually land in a separate chart.
\end{enumerate}

This algorithm is fast and produces charts that are reasonably efficient for packing (low count, non-complex boundaries). Figure~\ref{fig:atlas} shows a few typical examples.

\paragraph{Texture filter padding:}
When using mipmapping, filtering kernels span multiple consecutive pixels in a texture.
We therefore add a few pixel thick pad around each chart. 
We either copy redundant pixels from neighboring charts, or use isotropic diffusion at step-edges where pixels do not have neighbors across the chart boundary (dark/light red in Figure~\ref{fig:atlas}, respectively).

\paragraph{Macroblock padding:}
Another possible source of color bleeding is lossy image compression.
We encode textures with JPEG for transmission, which operates on non-overlapping $16 \times 16$ pixel macroblocks.
To avoid bleeding we smoothly inpaint any block that is overlapped by the chart with yet another round of isotropic diffusion (green pixels in Figure~\ref{fig:atlas}a).
Interestingly, this also reduces the encoded texture size by almost 40\% compared to solid color fill, \rev{because the step edges are pushed from the chart boundaries to macroblock boundaries where they become ``invisible'' for the JPEG encoder}.

\paragraph{Packing:}
Finally, we pack the padded charts into a single atlas image, so the whole mesh can be rendered as a single unit.
We use a simple tree-based bin packing algorithm\footnote{\url{http://blackpawn.com/texts/lightmaps/}}.
Figure~\ref{fig:atlas}c shows the complete atlas for the 3D photo in Figure~\ref{fig:teaser}.


\subsubsection{Meshing}
\label{sec:mesh}

In the final stage of our algorithm, we create a triangle mesh that is textured using the atlas from the previous section.
A dense mesh with micro-triangles can be trivially constructed from the LDI by replacing pixels with vertices and connections with triangles.
However, this would be prohibitively large to render, store, and transmit over a network.

Simplification algorithms for converting detailed meshes into similar versions with fewer triangles are a long-studied area of computer graphics.
However, even advanced algorithms are relatively slow when applied to such large meshes.

Therefore, we designed a custom algorithm that constructs a simplified mesh directly.
It exploits the 2.5D structure of our representation, by operating in the 2D texture atlas domain: simplifying and triangulating the chart polygons first in 2D, and then lifting them to 3D later.

We start by converting the outline of each chart into a detailed 2D polygon, placing vertices at the corners between pixels (Figure~\ref{fig:meshing}a).
Next, we simplify the polygon using the Douglas-Peucker algorithm \shortcite{Douglas1973} (Figure~\ref{fig:meshing}b).
Most charts share some parts of their boundary with other charts that are placed elsewhere in the atlas (e.g., light red padding pixels in Figure~\ref{fig:atlas}a).
We are careful to simplify these shared boundaries in the exact same way, so they are guaranteed to fit together when re-assembling the charts.

Now we are ready to triangulate the chart interiors.
It is useful to distribute internal vertices to be able to reproduce depth variations and achieve more regular triangle shapes.
We considered using adaptive sampling algorithms but found their degree of sophistication unnecessary, since all major depth discontinuities are already captured at chart boundaries, and the remaining parts are relatively smooth in depth.
\rev{We therefore simply generate strips of vertical ``stud'' polylines with evenly spaced interior vertices (Figure~\ref{fig:meshing}c).
The studs are placed as evenly as possible, given the constraint that they have to start and end on chart boundary vertices.}
We triangulate the composite polygon using a fast plane-sweep algorithm \cite{Berg2008} (Figure~\ref{fig:meshing}d).

Having obtained a 2D triangulation, we now simply lift it to 3D by projecting every vertex along its corresponding ray according to its depth (Figure~\ref{fig:meshing}e).
This 3D triangle mesh, together with the atlas from the previous section, is our final representation.

%% file: figures/meshing.tex
\newlength\fma
\setlength\fma{3.4cm}

\begin{figure*}
  \centering%
  \jsubfig{\includegraphics[height=\fma]{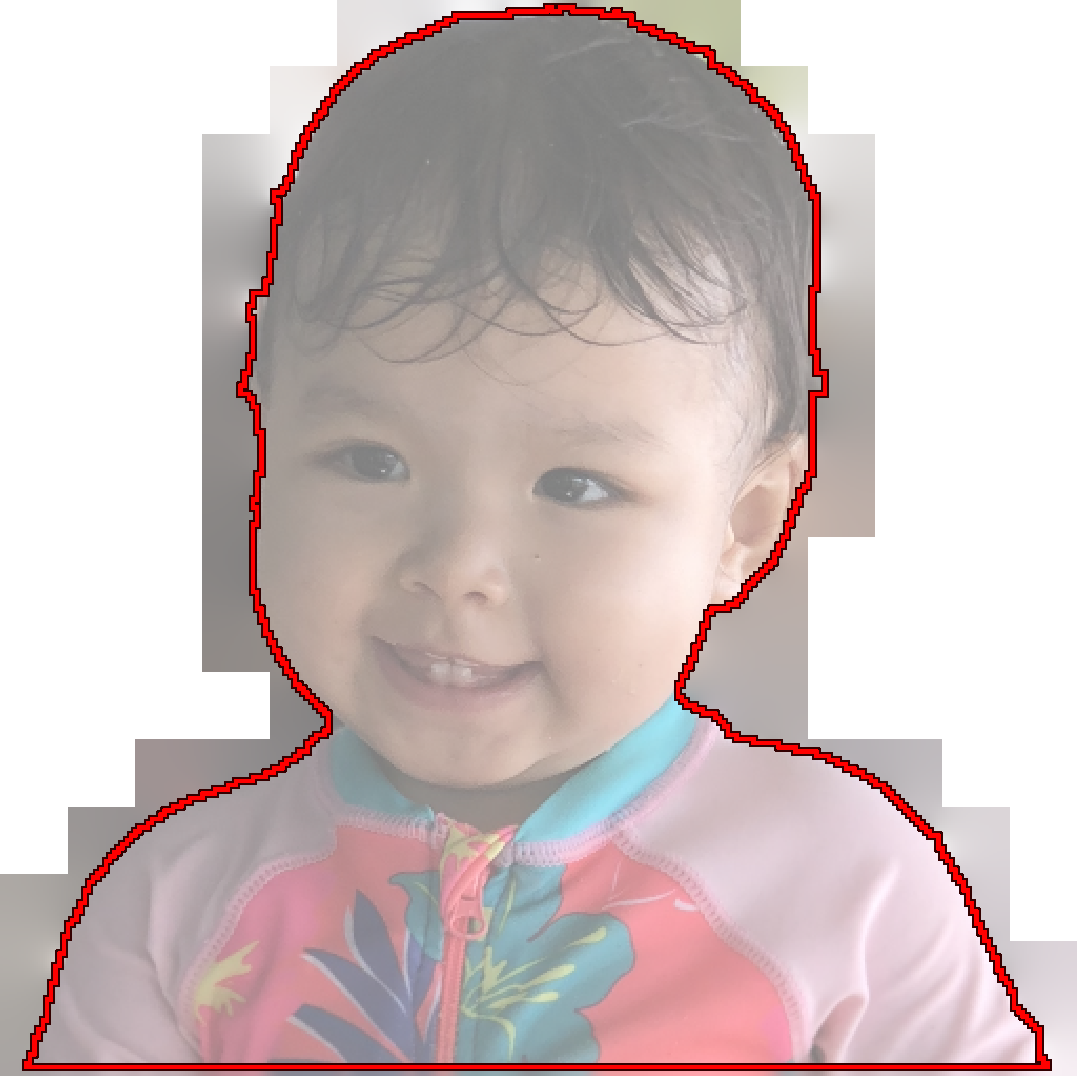}}{(a) Detailed polygon}%
	\hfill%
  \jsubfig{\includegraphics[height=\fma]{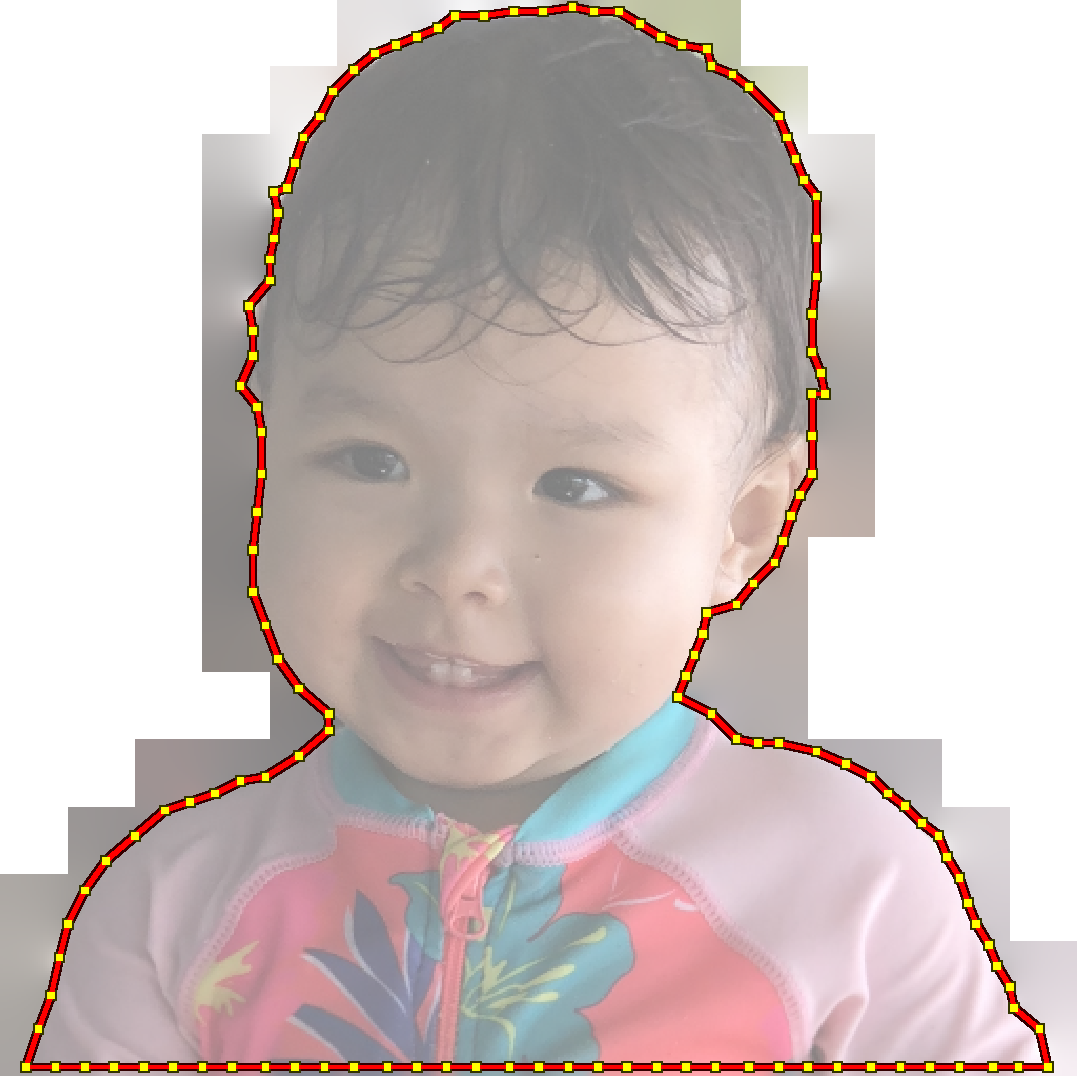}}{(b) Simplified polygon}%
	\hfill%
  \jsubfig{\includegraphics[height=\fma]{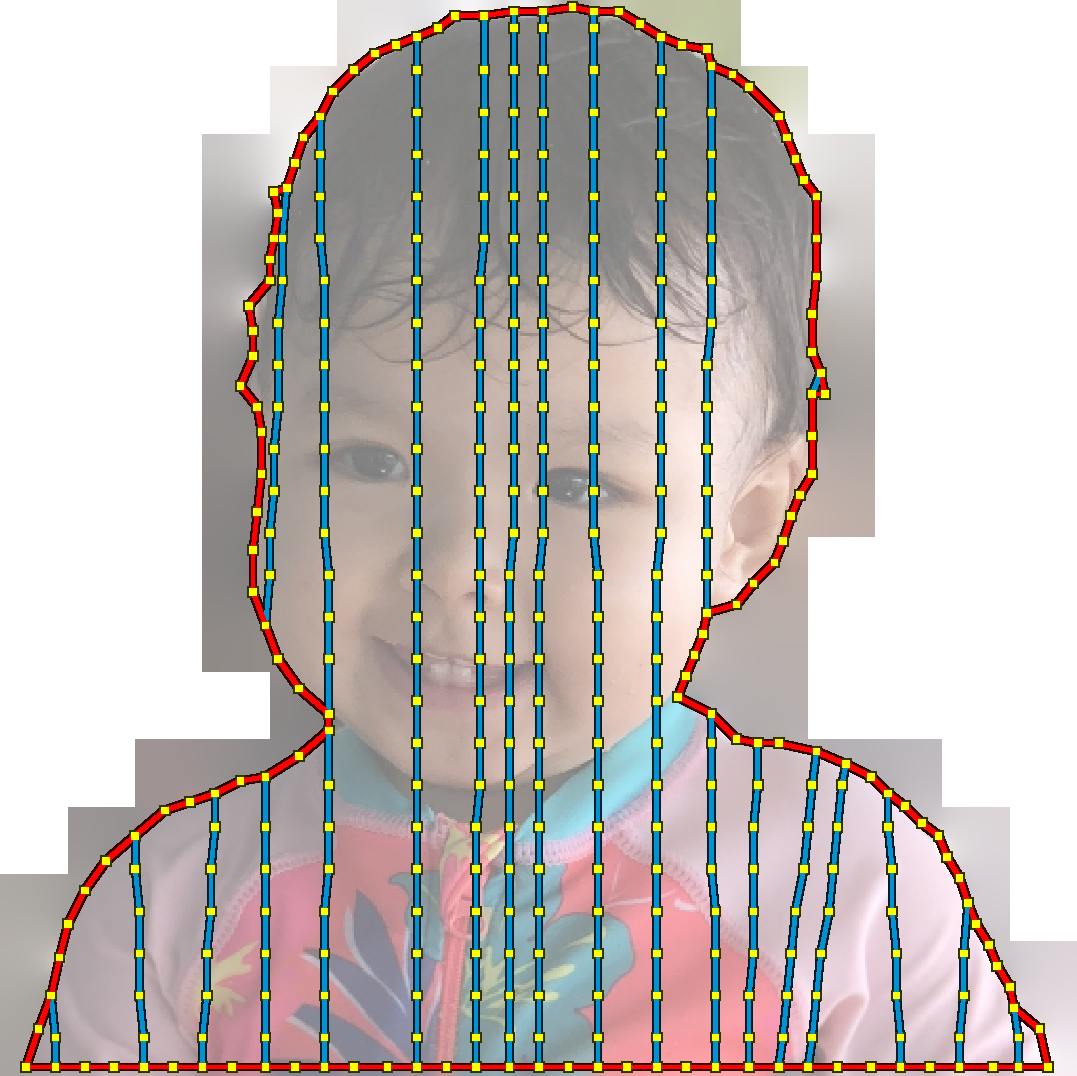}}{(c) Interior vertices}%
	\hfill%
  \jsubfig{\includegraphics[height=\fma]{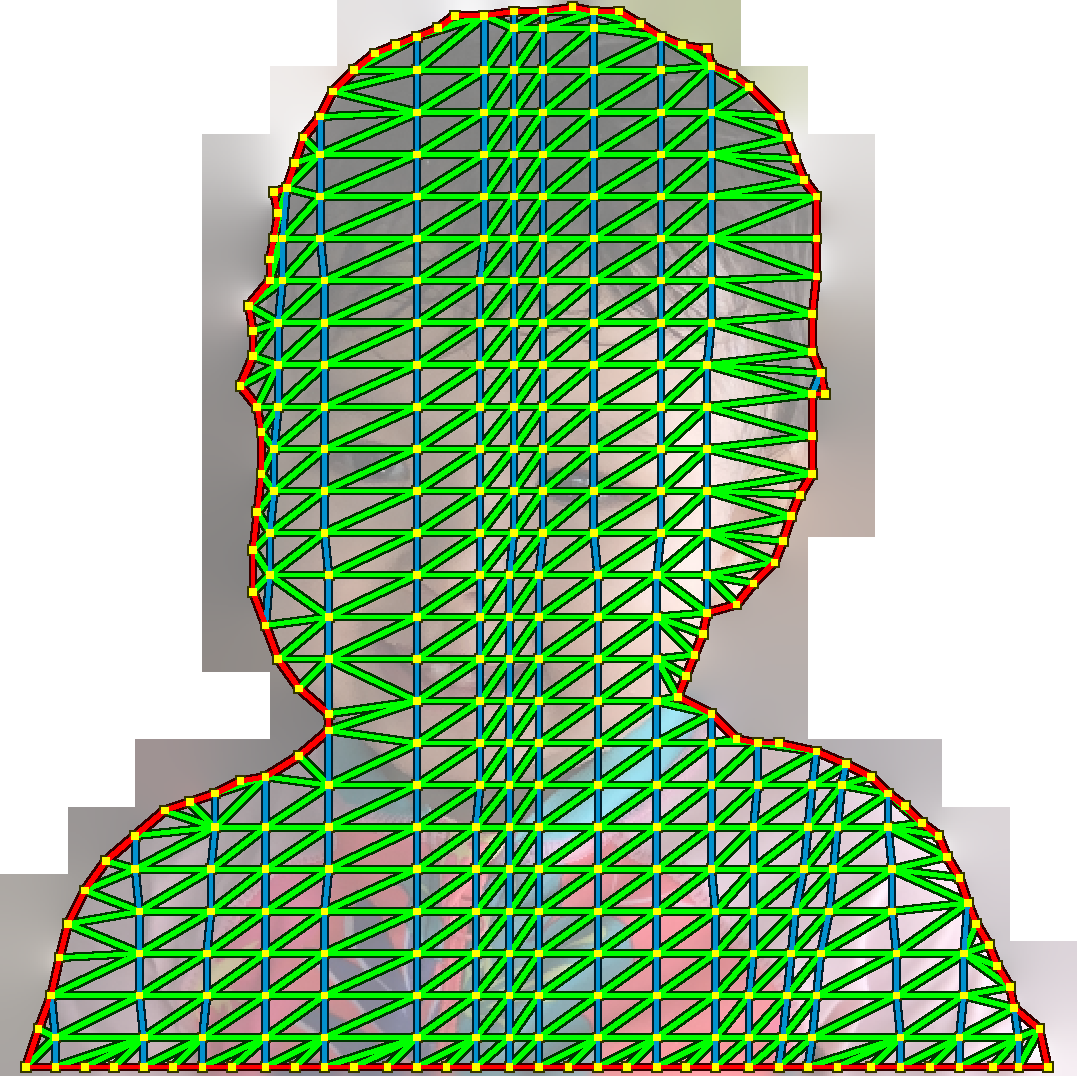}}{(d) 2D triangulation}%
	\hfill%
  \jsubfig{\includegraphics[height=\fma]{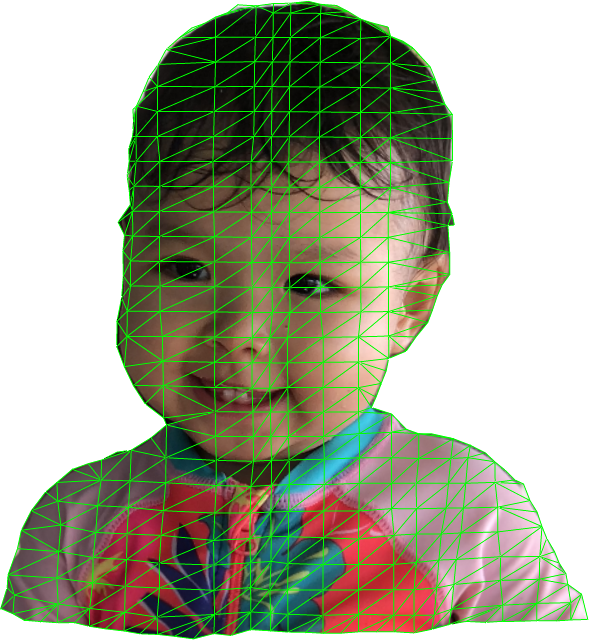}}{(d) Lifted to 3D}%
	\vspace{-3mm}\\%
	\def\shortdesc{Directly constructing a simplified triangle mesh in the 2D atlas domain.}
	\def\longdesc{
\shortdesc
(a) Detailed polygon from the outline of a single chart.
(b) Simplified chart polygon. Adjacent charts are simplified identically to guarantee a tight fit.
(c) Added interior vertices to represent depth variation and achieve more regular triangle shapes.
(d) 2D triangulation.
(e) Lifting the mesh to 3D by projecting vertices along their corresponding rays according to their depth.
}
	\caption{\longdesc}
	\Description[\shortdesc]{\longdesc}
	\undef\shortdesc
	\undef\longdesc
  \label{fig:meshing}
\end{figure*}

%% file: tex/Viewing.tex
\section{Viewing 3D Photos}
\label{sec:viewing}

\input{figures/viewing.tex}

Without motion, a 3D photo is just a 2D photo.
Fully experiencing the 3D format requires moving the virtual viewpoint to recreate the parallax one would see in the real world. We have designed interfaces for both mobile devices and desktop browsers, as well as for head-mounted VR displays, where we also leverage stereo viewing.

\subsection{Mobile and Browser}

On mobile devices, there are a number of possible affordances that can be mapped to virtual camera motion. These include scrolling in the application interface, using the device's IMUs such as the gyros to detect rotation, and using touch to manually rotate the view. 

After considerable user testing, mapping scrolling behavior to both vertical rotation (about the horizontal axis) as well as dollying in and out (translation along the ``z'' axis) emerged as the best set of control interactions. This gives the illusion while scrolling through a vertical feed that the viewing point moves up and into the scene. We also added a small bit of horizontal rotation (about the vertical axis) mapped to scrolling.
Furthermore, we add additional rotation to the virtual camera based on rotation of the device detected by gyros (see Figure~\ref{fig:viewing}).
In a web browser, we substitute mouse motion for gyro rotation.

\subsection{In Virtual Reality}

In VR, we have the advantage of being able to produce two offset images, one for each eye, to enable binocular stereo.
This creates a stronger feeling of immersion.
3D photos are currently the only \rev{photographic} user-generated content in VR that makes use of all degrees of freedom in this medium. 

We use threeJS (a Javascript 3D library) to render the scene to a WebGL context, and we use WebVR to render this context to a VR Device. The renderer queries the device parameters (eye buffer size and transforms), applying the information separately for the left and right eye views to produce a stereo image.

In addition to stereo, we map head motion directly to virtual camera motion. In 6-DOF headsets, this is a one-to-one mapping. In 3-DOF (rotation only), we mimic head translation from rotation around the neck since rotating the head to the left, for example, also translates the eyes leftward.

We create a frame around the model to hide the outer boundary of the photo.
The result appears like a 3D model viewed through 
a 2D frame. 
Since the quality of the 3D photo display degrades when moving too far away from the original viewpoint, we constrain the viewing angles and fade the model out if there is too much head motion.


%% file: figures/viewing.tex
\begin{figure}
  \centering%
    \includegraphics[width=\columnwidth]{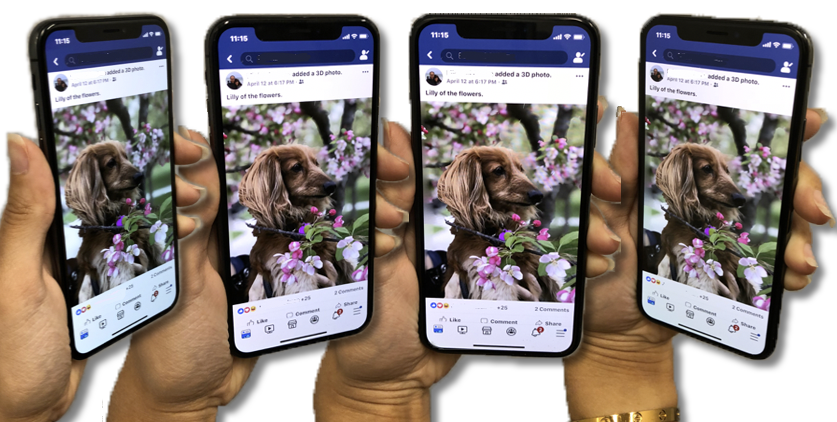}
	\def\desc{Rotating phone induces parallax through sensing from the gyro.}
	\caption{\desc}
	\Description[\desc]{\desc}
	\undef\desc
  \label{fig:viewing}
\end{figure}

%% file: tex/results.tex
\section{Results and Evaluation}
\label{sec:results}

\subsection{Results}

We have extensively tested the robustness of our system.
Early versions of the system have been deployed in a social media app, where they have been used over 100 million times, attesting to the quality and robustness of the algorithms.

Unlike most other view synthesis methods our systems takes only a \emph{single} color image as input.
We can therefore apply it to any pre-existing image.
In the supplementary video we show results on a wide range of historically significant photographs.
We also show a large variety of results on snapshots.

\subsection{Code}

Pretrained models of our depth estimation network and inpainting networks are publicly available at the project page.

\subsection{Performance}

The table below breaks out the runtime of our algorithm stages on a typical image.
We measured these numbers on an iPhone 11 Pro on six randomly selected $1152 \times 1536$ images.
Depth is estimated at $288 \times 384$ resolution in 230ms.
We report the median time for each stage.

\begin{center}
\begin{tabular}{l@{\hspace{1mm}}c@{\hspace{2mm}}c@{\hspace{4mm}}c@{\hspace{5mm}}c}
\textbf{Algorithm stage} & \textbf{Mean Runtime}\\
\hline
Depth estimation & 230ms\\
Depth filter & 51ms\\
Connecting components  & 12ms\\
Occluded geometry & 31ms\\
Color inpainting & 540ms\\
Texture chart generation & 72ms\\
Texture chart padding & 151ms\\
Meshing & 11ms\\
\hline
\textbf{Total} & 1098ms\\
\end{tabular}
\end{center}

We store the final textured mesh in a GL Transmission Format (glTF) container for transmission.
This representation can be rendered practically on device using standard graphics engines as discussed in Section \ref{sec:viewing}. 
The final size of the textured mesh representation is typically around 300-500kb for an input image of size $1152 \times 1536$.

An important advantage of our representation is that it is uses GPU memory efficiently.
While an MPI stores a stack of full-sized images, the texture atlas only represents surfaces that are actually used.

\subsection{Depth Estimation}
\label{sec:depth_eval}

\input{figures/depth_eval.tex}

We quantitatively compare our optimized depth estimation network against several state-of-the-baselines methods in Table~\ref{tab:depth_eval}.
For most methods the authors only provided fine-tuned models and no training code.
We list for each method the datasets it was trained with.
In the training data column RW refers to ReDWeb~\cite{Xian2018}, MD to MegaDepth~\cite{Li2018}, MV to Stereo Movies~\cite{Ranftl2019}, DL to DIML Indoor~\cite{Kim2018}, K to KITTI~\cite{Menze2015}, KB to Ken Burns~\cite{Niklaus2019}, CS to Cityscapes~\cite{Cordts2016}, WSVD~\cite{Wang2019}, PBRS~\cite{Zhang2016}, and NYUv2~\cite{Silberman2012}.
\rev{3DP refers to a proprietary dataset of 2.0M iPhone dual-camera images of a wide variety of scenes.}
A $\rightarrow$ B indicates that a model was pretrained on A and fine-tuned on B.

We compare against Midas~\cite{Ranftl2019} (versions 1 and 2 released in June 2019 and December 2019, 
 respectively), Monodepth2~\cite{Godard2019}, SharpNet~\cite{Ramamonjisoa2019}, MegaDepth~\cite{Li2018}, Ken Burns~\cite{Niklaus2019}, and PyD-Net~\cite{Poggi2018}.

Each method has a preferred resolution at which it performs best.
These numbers are either explicitly listed in the respective papers or the author-provided code resizes inputs to the specific resolution.
Also, different methods have different alignment requirements (e.g., width/height must be a multiple of 16).
We list these details in the supplementary document, but briefly: all methods, except \rev{Monodepth2, Ken Burns, and PyD-Net} resize the input image so the long dimension is 384 and the other dimension is resized to preserve the aspect ratio.
Ken Burns \rev{and PyD-Net resize} the long dimension to \rev{1024 and 512, respectively}, and Monodepth2 uses a fixed $1024 \times 320$ aspect ratio.
For this evaluation we resize the input image to each algorithm's preferred resolution, and then resize the result to 384 pixels at which we compare against GT.
We evaluate models on the MegaDepth test split \cite{Li2018} as well as the entire ReDWeb dataset \cite{Wang2019}, and report standard metrics.
In the supplementary document we provide a larger number of standard metrics.
For the Midas networks we omit the ReDWeb numbers because it was trained on this dataset, and ReDWeb does not provide a test split.

\rev{
We evaluate four versions of our depth network:
\begin{description}
\item[Baseline:] refers to a manually crafted architecture, as described in Section~\ref{sec:depth}.
\item[AS + no-quant:] refers to an optimized architecture with float32 operators (i.e., no quantization).
\item[AS + quant:] refers to the optimized architecture with quantization. This is our full model.
\item[AS + quant, MD + 3DP:] for completeness we list another snapshot that was trained with a proprietary dataset of 2.0M dual-camera images.
\end{description}
}

We evaluate the performance on an example image of dimensions 384$\times$288.
We first report the FLOP count of the model, computed analytically from the network schematics.
Because FLOP counts do not always accurately reflect latency, we make runtime measurements on a mobile device.
At the same time, we measure peak memory consumption during the inference.
All models were run on an iPhone 11 Pro.
We ran the models on the device as follows.
All models came in PyTorch\footnote{\url{https://pytorch.org/}} format \rev{(except PyD-Net)}.
We converted them to Caffe2\footnote{\url{https://caffe2.ai/}} using ONNX\footnote{\url{https://onnx.ai/}}, because of Caffe2's mobile capabilities (Caffe2go).
We optimized the memory usage with the Caffe2 Memonger module.
\rev{Because our scripts did not work on the PyD-Net tensorflow model we omit it from the performance evaluation.}
Then we measured the peak memory consumption by hooking the Caffe2 allocator function and keeping track of the maximum total allocation during the network run.
Only Midas v1, Monodepth2, and our models were able to run on device, the other ones failed due to insufficient memory.
\rev{Both models have footprints that are more than an order of magnitude larger than ours.}

Finally, we provide details about the model size.
We list the number of float32 and int8 parameters for each model as well as the total model size in MiB, with smaller models being more amendable to mobile download. 
While our method does not perform best in terms of quality compared to significantly higher number of parameter state of the art models, it is competitive and the quality is sufficient for our intended application, as demonstrated by hundreds of results shown in the supplemental video.
The main advantages of our model is that its size is significantly smaller also resulting in significantly reduced computation compared to the state of the art models.
The enables depth estimation even on older phone hardware.

\subsection{Inpainting}

\input{figures/inpaint_eval_repro.tex}
\input{figures/ldi_vs_2d_inpainting.tex}
\input{figures/inpaint_eval_quant.tex}

We quantitatively evaluate inpainting on the ReDWeb dataset~\cite{Xian2018}, because it has dense depth supervision. In order to evaluate inpainting we follow this procedure:
\begin{itemize}
    \item For each image in the dataset we lift input image to single-layer LDI (i.e., no extending) and use micro-polygons. That is, we capture all detail, but we don't hallucinate any new details. (Figure~\ref{fig:ip_eproc}a).
    \item Then we render the image from a canonical viewpoint and use depth peeling to obtain an LDI with known colors at \emph{all} layers, not just the first one (Figure~\ref{fig:ip_eproc}b--c show the first two layers).
    \item We consider all layers except the first one as unknown and inpaint them (Figure~\ref{fig:ip_eproc}d). 
    \item Finally, we reproject the inpainted LDI back to the original view (Figure~\ref{fig:ip_eproc}e). This is useful, because in this view \emph{all} inpainted pixels (from any LDI layer) are visible, and because it is a normal rendered image we can use any image-space metric. 
\end{itemize}

In the  ``Quality (LDI)'' column in Table~\ref{tab:inpaint_eval} we report a quality loss computed on the LDI (i.e., between Figures~\ref{fig:ip_eproc}c and \ref{fig:ip_eproc}d).
In the ``Quality (reprojected)'' column in Table~\ref{tab:inpaint_eval} we report PSNR and SSIM metrics. Since SSIM and PSNR evaluate for reconstruction error, we also include the LPIPS metric~\cite{Zhang2018} to better evaluate the perceptual similarity of the inpainted image compared to the ground truth. 

We compare our optimized model against original full size PartialConv model. We also compare against using both models applied in regular screen space inpainting. Fig. \ref{fig:ldiv2d} illustrates the significant artifacts on the edges when naively using regular screen space inpainting. 




\rev{
\subsection{End-to-end View Synthesis}

In the supplementary material we provide a qualitative comparison to the ``3D Ken Burns Effect''~\cite{Niklaus2019}.
Note that the output of that system is a video showing a linear camera trajectory, and their inpainting is optimized solely for viewpoints along that trajectory.
In contrast, the output of our system is a mesh that is suitable for rendering from any viewpoint near the point of capture.
}

\subsection{Limitations}

As with any computer vision method, our algorithm does not always work perfectly.
The depth estimation degrades in situations that are not well represented in the training data.
An inherent limitation of the depth representation is that there is only one depth value per pixel in the input; semi-transparent surfaces or participating media (e.g., fog or smoke) are not well represented.
We thus see a number of cases where the resulting 3D photo suffers from bad depth values. Nevertheless, most scene captures do result in successful 3D photos.
The sets of images in the two ``results'' parts in the supplemental video, were only selected for content before applying our algorithm.
We did not remove any failure cases based on processing.
Therefore, you can see some artifacts if examined closely.
They thus provide an idea of the success rate of the algorithm.


%% file: figures/depth_eval.tex
\begin{table*}[t]
\centering
\caption{\rev{
Quantitative evaluation our our depth estimation network.
The best performance in each column is set in \textbf{bold}, and the second best \underline{underscored}.
Note, that for the quality evaluation every network used its preferred resolution, while for the performance evaluation we used a fixed resolution of $384 \times 288$ for all networks.
Please refer to the text for a detailed explanation.
}}
\label{tab:depth_eval}
\begin{adjustbox}{max width=\textwidth}
\begin{tabular}{ll|S[table-format=1.3]S[table-format=1.3]S[table-format=1.3]|S[table-format=1.3]S[table-format=1.3]S[table-format=1.3]|S[table-format=2.1]S[table-format=1.2]S[table-format=3.1]|S[table-format=3.1]S[table-format=3.1]S[table-format=3.1]}\toprule
&& \multicolumn{3}{c|}{Quality (MegaDepth)} & \multicolumn{3}{c|}{Quality (ReDWeb)} & \multicolumn{3}{c|}{Performance} &  \multicolumn{3}{c}{Model footprint}\\
Method & Training data &
{$\delta\!<\!1.25\!\uparrow$} & {Abs rel$\downarrow$} & {RMSE$\downarrow$} &
{$\delta\!<\!1.25\!\uparrow$} & {Abs rel$\downarrow$} & {RMSE$\downarrow$} &
{FLOPs$\downarrow$} & {Runtime$\downarrow$} & {Peak mem.$\downarrow$} &
{float32} & {int8} & {Size$\downarrow$}\\
\midrule
Midas (v1) & \small RW, MD, MV &
\underline{0.955} & \underline{0.068} & 0.027 &
{-} & {-} & {-} &
\SI{33.2}{\giga\relax} & \SI{1.11}{\second} & \SI{453.7}{\mebi\byte} &
\SI{37.3}{\mega\relax} & {-} & \SI{142.4}{\mebi\byte}\\
Midas (v2) & \small RW, DL, MV, MD, WSVD &
\textbf{0.965} & \textbf{0.058} & \textbf{0.022} &
{-} & {-} & {-} &
\SI{72.3}{\giga\relax} & {-} & {-} &
\SI{104.0}{\mega\relax} & {-} & \SI{396.6}{\mebi\byte}\\
Monodepth2 & \small K &
0.845 & 0.145 & 0.049 &
0.350 & 4.368 & 0.176 &
\underline{\SI{6.7}{\giga\relax}} & \underline{\SI{0.26}{\second}} & \textbf{\SI{194.1}{\mebi\byte}} &
\SI{14.3}{\mega\relax} & {-} & \SI{54.6}{\mebi\byte}\\
SharpNet & \small PBRS $\rightarrow$ NYUv2 &
0.839 & 0.146 & 0.051 &
0.308 & 6.616 & 0.196 &
\SI{54.9}{\giga\relax} & {-} & {-} &
\SI{114.1}{\mega\relax} & {-} & \SI{435.1}{\mebi\byte}\\
MegaDepth & \small DIW $\rightarrow$ MD &
0.929 & 0.086 & 0.033 &
\underline{0.434} & 2.270 & \textbf{0.137} &
\SI{63.2}{\giga\relax} & {-} & {-} &
\SI{5.3}{\mega\relax} & {-} & \SI{20.4}{\mebi\byte}\\
Ken Burns & \small MD, NYUv2, KB &
0.948 & 0.070 & \underline{0.026} &
\textbf{0.438} & 2.968 & \underline{0.140} &
\SI{59.4}{\giga\relax} & {-} & {-} &
\SI{99.9}{\mega\relax} & {-} & \SI{381.0}{\mebi\byte}\\
\rev{PyD-Net} & \rev{\small CS $\rightarrow$ K} &
\rev{0.836} & \rev{0.148} & \rev{0.052} &
\rev{0.310} & \rev{5.218} & \rev{0.198} &
\rev{{-}} & \rev{{-}} & \rev{{-}} & 
\rev{\SI{2.0}{\mega\relax}} & \rev{{-}} & \rev{\underline{\SI{7.9}{\mebi\byte}}}\\
\midrule
\rev{Tiefenrausch \small{(baseline)}} & \rev{\small MD} &
\rev{0.942} & \rev{0.078} & \rev{0.031} &
\rev{0.383} & \rev{1.961} & \rev{0.156} &
\rev{\SI{18.9}{\giga\relax}} & \rev{{-}} & \rev{{-}} &
\rev{\SI{3.0}{\mega\relax}} & \rev{{-}} & \rev{\SI{11.4}{\mebi\byte}}\\
\rev{Tiefenrausch \small{(AS + no-quant)}} & \rev{\small MD} &
\rev{0.940} & \rev{0.080} & \rev{0.031} &
\rev{0.378} & \rev{1.987} & \rev{0.157} &
\rev{\textbf{\SI{6.4}{\giga\relax}}} & \rev{{-}} & \rev{{-}} &
\rev{\SI{3.5}{\mega\relax}} & \rev{{-}} & \rev{\SI{13.4}{\mebi\byte}}\\
Tiefenrausch \small{(AS + quant)} & \small MD &
0.941 & 0.079 & 0.031 &
0.382 & \underline{1.950} & 0.156 &
\textbf{\SI{6.4}{\giga\relax}} & \textbf{\SI{0.23}{\second}} & \underline{\SI{196.1}{\mebi\byte}} &
{-} & \SI{3.5}{\mega\relax} & \textbf{\SI{3.3}{\mebi\byte}}\\
\rev{Tiefenrausch \small{(AS + quant)}} & \rev{\small MD, 3DP} &
\rev{0.925} & \rev{0.090} & \rev{0.035} &
\rev{0.407} & \rev{\textbf{1.541}} & \rev{0.142} &
\rev{\textbf{\SI{6.4}{\giga\relax}}} & \rev{\textbf{\SI{0.23}{\second}}} & \rev{\underline{\SI{196.1}{\mebi\byte}}} &
\rev{{-}} & \rev{\SI{3.5}{\mega\relax}} & \rev{\textbf{\SI{3.3}{\mebi\byte}}}\\
\bottomrule
\end{tabular}
\ignorethis{
\begin{tabular}{ll|S[table-format=1.3]S[table-format=1.3]S[table-format=1.3]|S[table-format=1.3]S[table-format=1.3]S[table-format=1.3]|S[table-format=2.1]S[table-format=1.2]S[table-format=3.1]|S[table-format=3.1]S[table-format=3.1]S[table-format=3.1]}\toprule
&& \multicolumn{3}{c|}{Quality (MegaDepth)} & \multicolumn{3}{c|}{Quality (ReDWeb)} & \multicolumn{3}{c|}{Performance} &  \multicolumn{3}{c}{Model footprint}\\
Method & Training data &
{$\delta\!<\!1.25\!\uparrow$} & {Abs rel$\downarrow$} & {RMSE$\downarrow$} &
{$\delta\!<\!1.25\!\uparrow$} & {Abs rel$\downarrow$} & {RMSE$\downarrow$} &
{FLOPs$\downarrow$} & {Runtime$\downarrow$} & {Peak mem.$\downarrow$} & {float32} & {int8} & {Size$\downarrow$}\\
\midrule
%
Midas (v1) & \small RW, MD, MV & 0.955 & 0.068 & 0.027 & {-} & {-} & {-} & \SI{33.2}{\giga\relax} & \SI{1.11}{\second} & \SI{453.7}{\mebi\byte} & \SI{37.3}{\mega\relax} & {-} & \SI{142.4}{\mebi\byte}\\
\rowcolor{altcolor}
Midas (v2) & \small RW, DL, MV, MD, WSVD & 0.965 & 0.058 & 0.022 & {-} & {-} & {-} & \SI{72.3}{\giga\relax} & {-} & {-} & \SI{104.0}{\mega\relax} & {-} & \SI{396.6}{\mebi\byte}\\
Monodepth2 & \small K & 0.845 & 0.145 & 0.049 & 0.350 & 4.368 & 0.176 & \SI{6.7}{\giga\relax} & \SI{0.26}{\second} & \SI{194.1}{\mebi\byte} & \SI{14.3}{\mega\relax} & {-} & \SI{54.6}{\mebi\byte}\\
\rowcolor{altcolor}
SharpNet & \small PBRS $\rightarrow$ NYUv2 & 0.839 & 0.146 & 0.051 & 0.308 & 6.616 & 0.196 & \SI{54.9}{\giga\relax} & {-} & {-} & \SI{114.1}{\mega\relax} & {-} & \SI{435.1}{\mebi\byte}\\
MegaDepth & \small DIW $\rightarrow$ MD & 0.929 & 0.086 & 0.033 & 0.434 & 2.270 & 0.137 & \SI{63.2}{\giga\relax} & {-} & {-} & \SI{5.3}{\mega\relax} & {-} & \SI{20.4}{\mebi\byte}\\
\rowcolor{altcolor}
Ken Burns & \small MD, NYUv2, KB & 0.948 & 0.070 & 0.026 & 0.438 & 2.968 & 0.140 & \SI{59.4}{\giga\relax} & {-} & {-} & \SI{99.9}{\mega\relax} & {-} & \SI{381.0}{\mebi\byte}\\
\midrule
\rowcolor{ourcolor}
Tiefenrausch \small{(ours)} & \small MD & 0.941 & 0.079 & 0.031 & 0.382 & 1.950 & 0.156 & \SI{6.4}{\giga\relax} & \SI{0.23}{\second} & \SI{196.1}{\mebi\byte} & {-} & \SI{3.5}{\mega\relax} & \SI{3.3}{\mebi\byte}\\
\rowcolor{altcolor}
\bottomrule
\end{tabular}
\ignorethis}
\end{adjustbox}
\end{table*}

%% file: figures/inpaint_eval_repro.tex
\undef\hgap%
\newlength\height
\setlength\height{4.2cm}
\newlength\width
\setlength\width{2cm}
\newlength\hgap
\setlength\hgap{1mm}
\newlength\vneg
\setlength\vneg{-0.8mm}
\newcommand{\fig}[1]{\includegraphics[height=\height, trim=0 150 0 0, clip]{#1}}
\begin{figure*}[t]
\centering%
\jsubfig{\fig{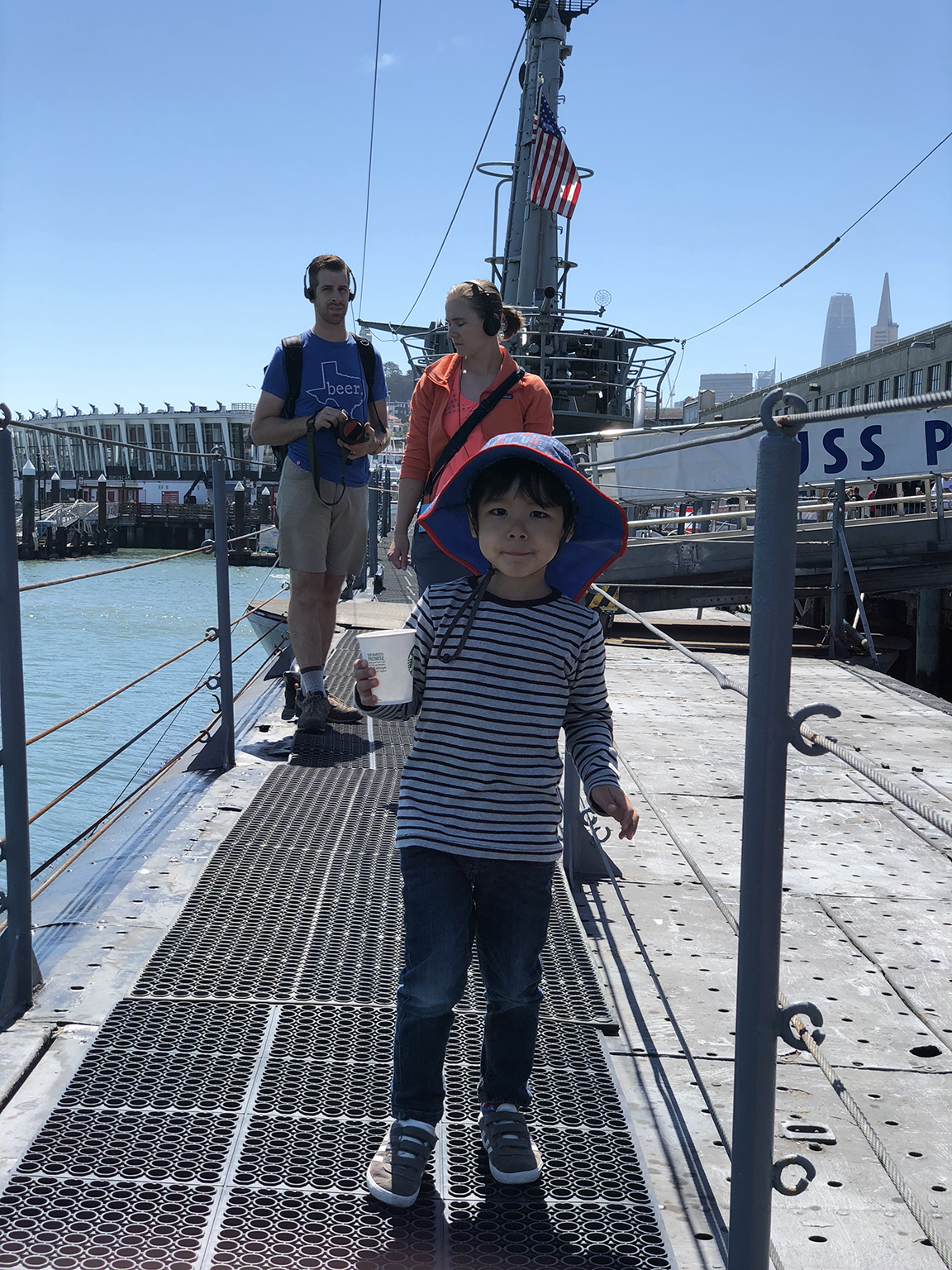}}{\vspace{\vneg}\small (a) Original view, ground truth}%
\hfill%
$\underbracket[0.5pt][2.0mm]{%
\jsubfig{\fbox{\fig{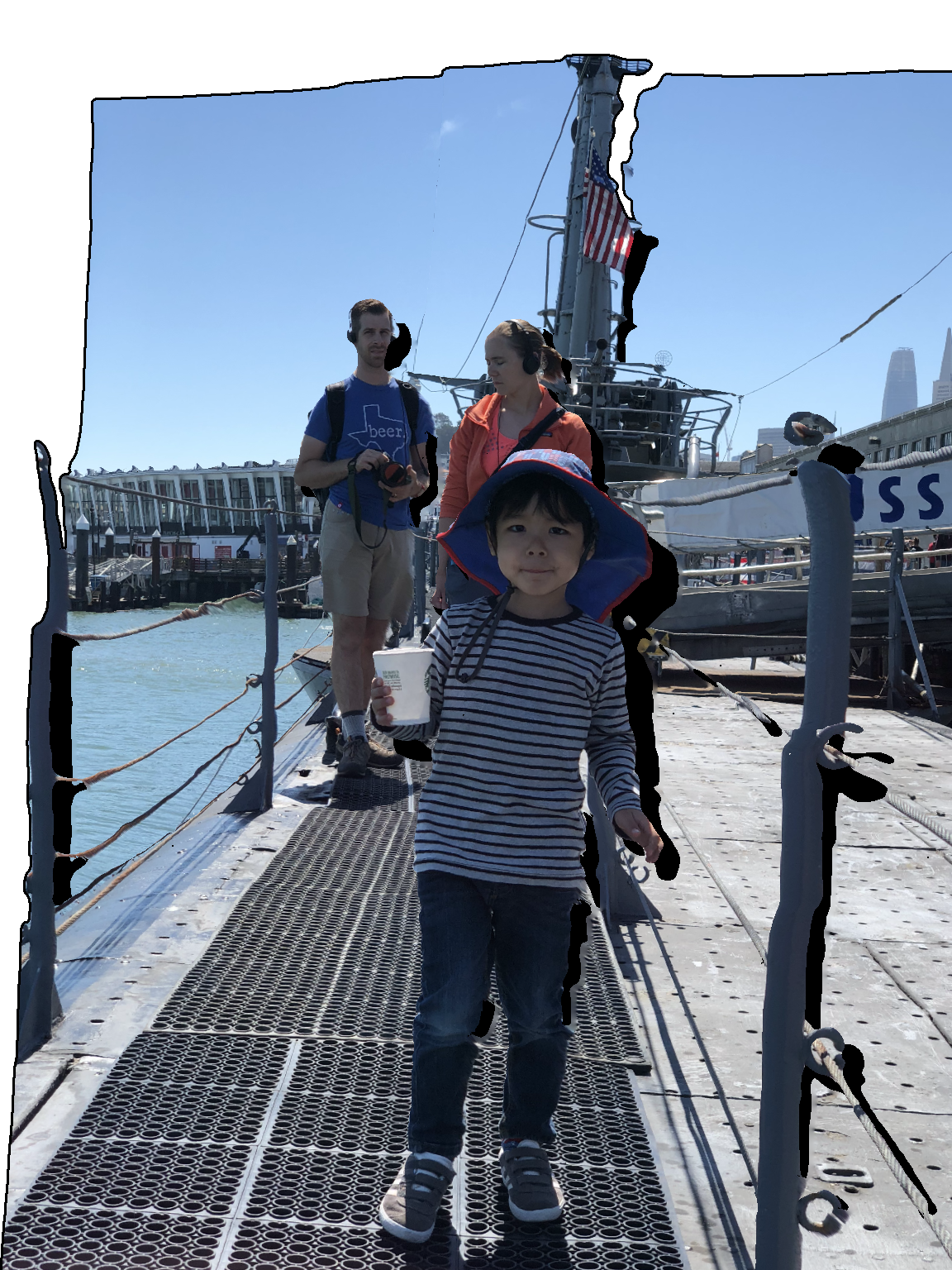}}}{\vspace{\vneg}\small (b) First layer, ground truth}%
\hspace{\hgap}
\jsubfig{\fbox{\fig{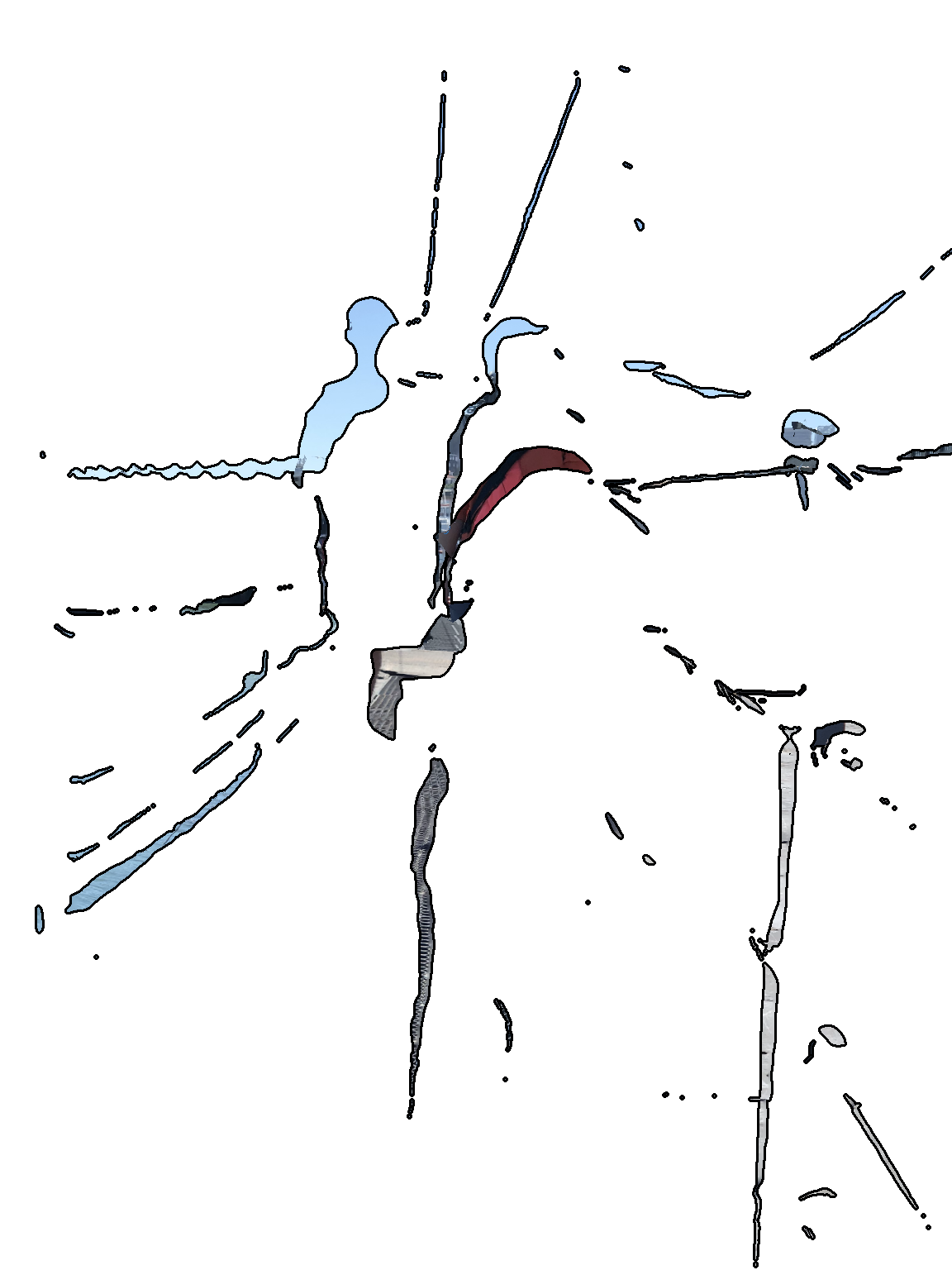}}}{\vspace{\vneg}\small (c) Second layer, ground truth}%
\hspace{\hgap}
\jsubfig{\fbox{\fig{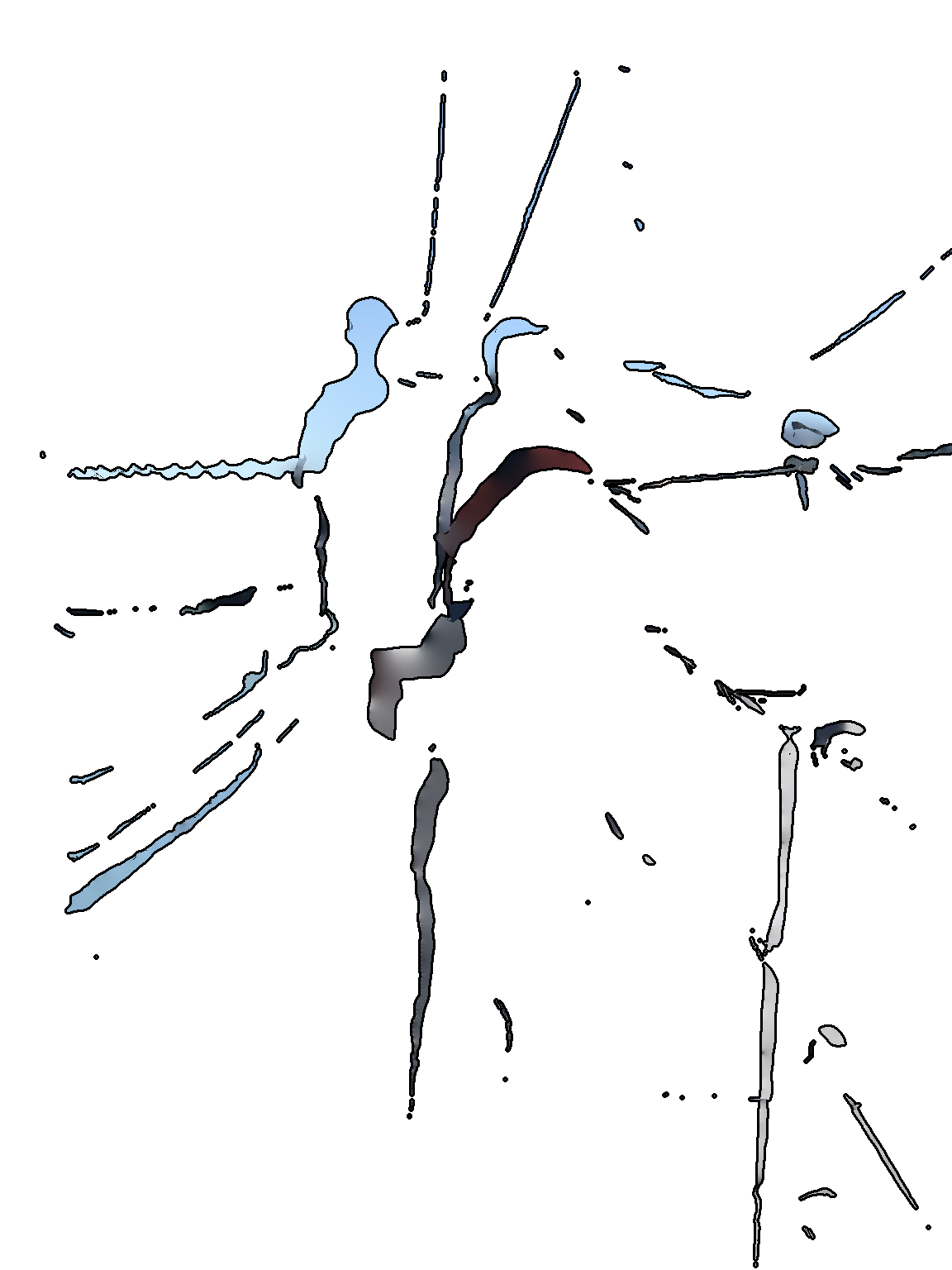}}}{\vspace{\vneg}\small (d) Second layer, Inpainted}%
}_{\substack{\vspace{-3.5mm}\\\colorbox{white}{~~\small Novel view~~}}}$%
\hfill%
\jsubfig{\fbox{\fig{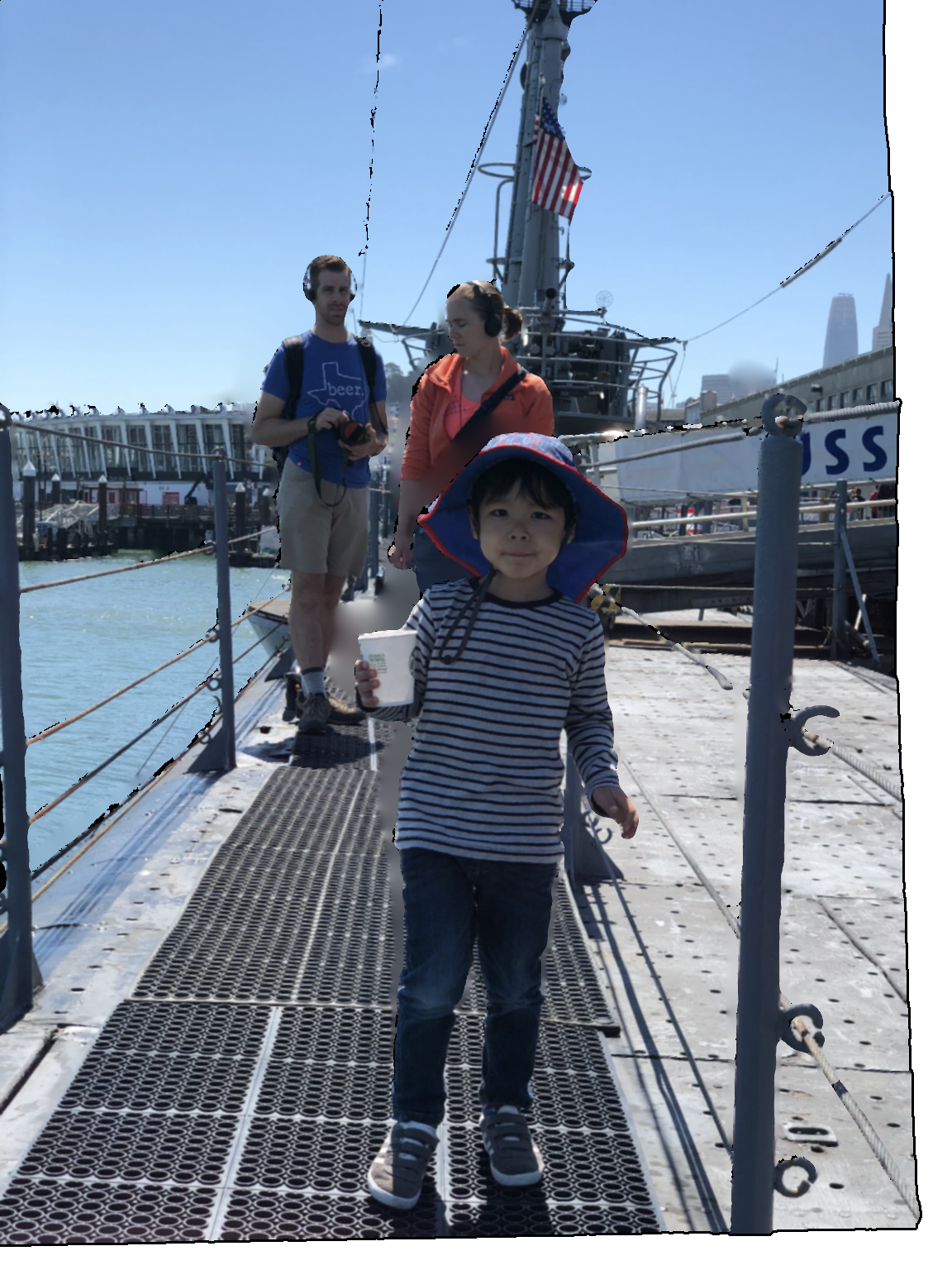}}}{\vspace{\vneg}\small (e) Original view, inpainted}%
\vspace{-3mm}
\caption{Inpainting evaluation.
In order to evaluate inpainting we follow this procedure:
left input image to single-layer LDI (i.e., no extending) 
}
\label{fig:ip_eproc}
\end{figure*}

\undef\height
\undef\width
\undef\hgap
\undef\vneg
\undef\fig

%% file: figures/ldi_vs_2d_inpainting.tex
\undef\hgap%
\newlength\height
\setlength\height{4.8cm}
\newlength\width
\setlength\width{2cm}
\newlength\hgap
\setlength\hgap{1mm}
\newlength\vneg
\setlength\vneg{-0.8mm}
\newcommand{\fig}[1]{\includegraphics[height=\height, trim=0 150 0 0, clip]{#1}}
\begin{figure*}[t]
\centering%
\jsubfig{\fig{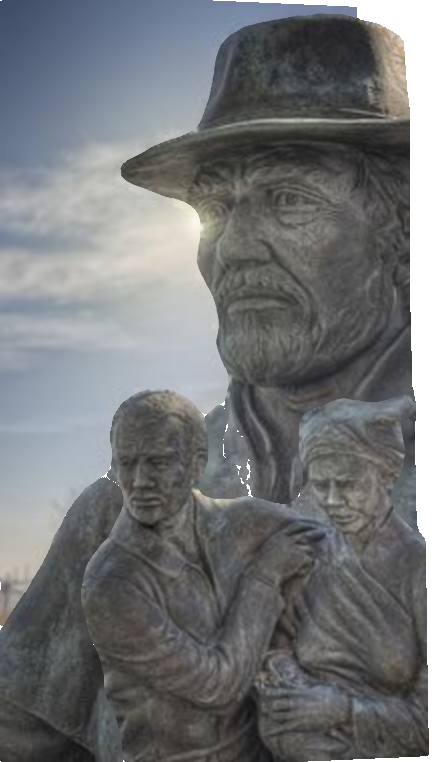}}{\vspace{\vneg}\small
(a) Ground truth}%
\hfill%
\jsubfig{\fig{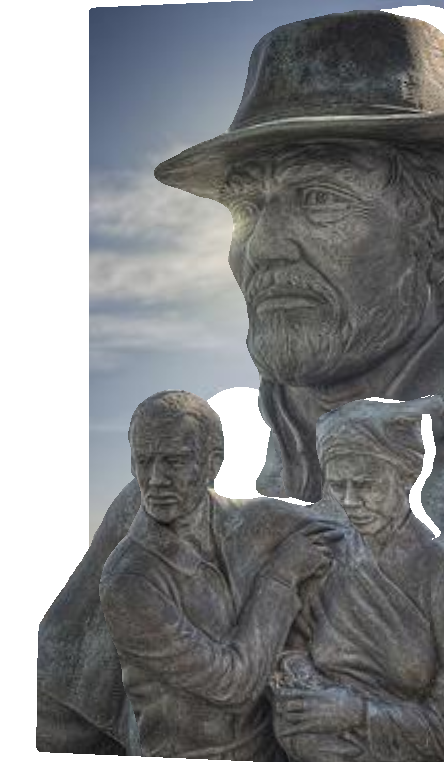}}{\vspace{\vneg}\small
(b) Wrapped view}%
\hfill%
\jsubfig{\fig{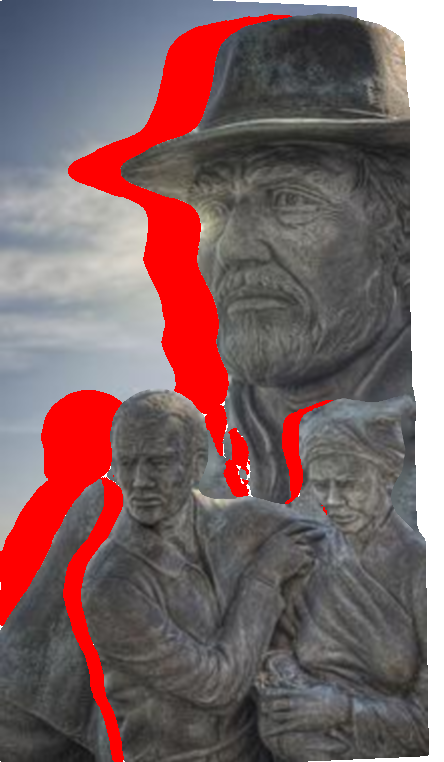}}{\vspace{\vneg}\small
(c) Masked input}%
\hfill%
\jsubfig{\fig{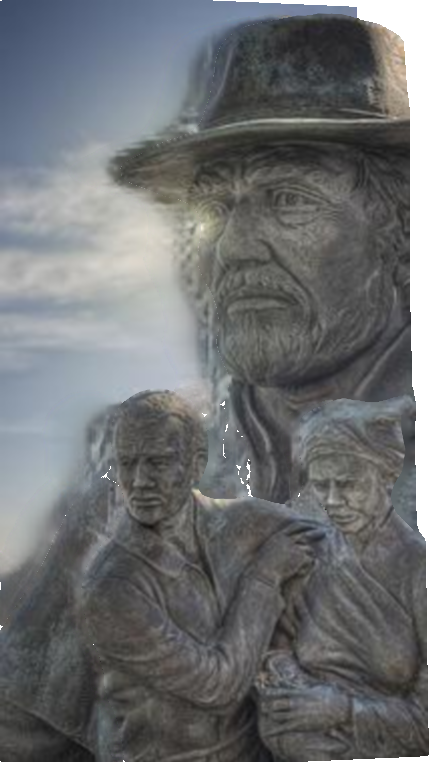}}{\vspace{\vneg}\small
(d) Farbrausch screen space inpainting}%
\hfill%
\jsubfig{\fig{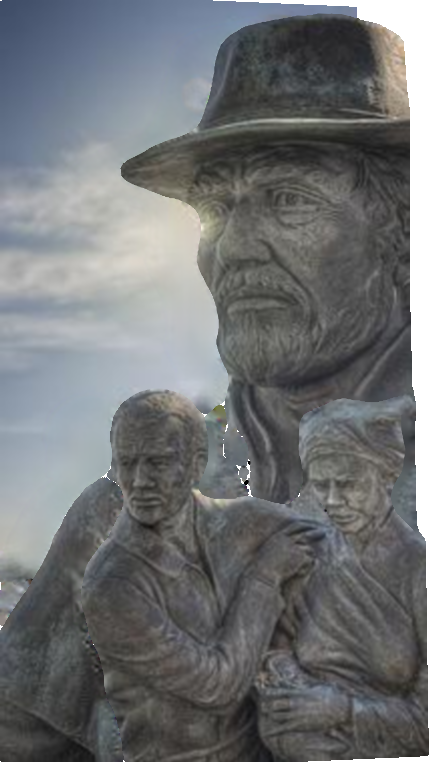}}{\vspace{\vneg}\small (e) Farbrausch LDI inpainting}%
\vspace{-3mm}
\caption{Comparing screen space inpainting to our LDI space inpainting using the same network (Farbrausch) trained on 2D images.
Source image \copyright~flickr user -ytf-, \url{https://www.flickr.com/photos/ytf/6837709270/}, used with permission.
}
\label{fig:ldiv2d}
\end{figure*}

\undef\height
\undef\width
\undef\hgap
\undef\vneg
\undef\fig

%% file: figures/inpaint_eval_quant.tex
\begin{table*}
\caption{\rev{Inpainting Evaluation.
The best performance in each column is set in \textbf{bold}, and the second best \underline{underscored}.
Note, FLOP count depends on the input size, and in the case of LDI this number is variable depending upon the geometric complexity;
we report FLOP counts for screen space inpainting with an image resolution of $512 \x 512$}.
}
\label{tab:inpaint_eval}
\definecolor{oddcolor}{gray}{0.9}
\definecolor{evncolor}{gray}{1.0}
\begin{tabular}{l|c|ccc|c|cc}
\toprule
\rowcolor{evncolor}
& Quality (LDI) &
\multicolumn{3}{c|}{Quality (reprojected)} &
\multicolumn{1}{c|}{Performance} &
\multicolumn{2}{c}{Model footprint}\\
Method &
PSNR$\uparrow$ & PSNR$\uparrow$ & SSIM$\uparrow$ & LPIPS$\downarrow$ &
FLOPs$\downarrow$ &
float32$\downarrow$ & Caffe2 Size$\downarrow$\\
\midrule
Farbrausch &
\textbf{33.852} & \textbf{34.126} & \underline{0.9829} & \underline{0.0232} &
- & 
\textbf{\SI{0.37}{\mega\relax}} & \textbf{\SI{1.9}{\mebi\byte}} \\
Partial Convolution &
\underline{33.795} & \underline{34.001} & \textbf{0.9832} & \textbf{0.0224} &
- & 
\underline{\SI{32.85}{\mega\relax}} & \underline{\SI{164.4}{\mebi\byte}} \\
\midrule
Farbrausch \small (screen space) & 
- & 32.0211 & 0.9784 & 0.0325 &
\textbf{\SI{2.56}{\giga\relax}} & 
\textbf{\SI{0.37}{\mega\relax}} & \textbf{\SI{1.9}{\mebi\byte}} \\
Partial Convolution \small (screen space) & 
- & 33.225 & 0.9807 & 0.0280 &
\underline{\SI{37.97}{\giga\relax}} & 
\underline{\SI{32.85}{\mega\relax}} & \underline{\SI{164.4}{\mebi\byte}} \\
\bottomrule
\end{tabular}
\ignorethis{
\begin{adjustbox}{max width=\textwidth}
\begin{tabular}{l|c|ccc|c|cc}
\toprule
& Quality (LDI) &
\multicolumn{1}{c|}{Performance} &
\multicolumn{2}{c}{Model footprint}\\
Method &
PSNR$\uparrow$ & PSNR$\uparrow$ & SSIM$\uparrow$ & LPIPS$\downarrow$ &
FLOPs*$\downarrow$ &
float32$\downarrow$ & Caffe2 Size$\downarrow$\\
\midrule
\rowcolor{ourcolor}
Farbrausch &
33.852 & 34.126 & 0.9829 & 0.0232 &
- & 
\SI{0.37}{\mega\relax} & \SI{1.9}{\mebi\byte} \\
Partial Convolution &
33.795 & 34.001 & 0.9832 & 0.0224 &
- & 
\SI{32.85}{\mega\relax} & \SI{164.4}{\mebi\byte} \\
%
%
\midrule
Farbrausch \small (screen space) & 
- & 32.0211 & 0.9784 & 0.0325 &
\SI{2.56}{\giga\relax} & 
\SI{0.37}{\mega\relax} & \SI{1.9}{\mebi\byte} \\
\rowcolor{altcolor}
Partial Convolution \small (screen space) & 
- & 33.225 & 0.9807 & 0.0280 &
\SI{37.97}{\giga\relax} & 
\SI{32.85}{\mega\relax} & \SI{164.4}{\mebi\byte} \\
%
%
\bottomrule
\end{tabular}
\end{adjustbox}
}
\end{table*}

%% file: tex/conclusions.tex
\section{Conclusions and future work}
\label{sec:conclusions}
In this work, we presented a new medium, a \emph{3D Photo}, and a system to produce them on any mobile device starting from a single image.
These 3D photos can be consumed on any mobile device as well through desktop browsers.  Scrolling, device motion, or mouse motion all induce virtual viewpoint change and thereby motion parallax.  3D Photos also are viewable in HMDs enabling stereoscopic viewing responsive to head motion.

Not only have we described the steps necessary to produce 3D Photos, but we've also presented advancements in optimizing depth and inpainting neural networks to run more efficiently on mobile devices.  These advancements can be used to improve fundamental algorithmic building blocks for Augmented Reality experiences.

There are many avenues for exploring human-in-the-loop creative expression with the 3D photo format. While this work shows how to auto-generate a 3D photo using real imagery, a future direction is to build out a rich set of creative tools to accommodate artistic intent.

\newcommand\blfootnote[1]{%
  \begingroup
  \renewcommand\thefootnote{}\footnote{#1}%
  \addtocounter{footnote}{-1}%
  \endgroup
}

\blfootnote{Images of individuals in this paper are used with permission.}